\documentclass[11pt]{article}

\usepackage[preprint]{acl}

\usepackage{times}
\usepackage{latexsym}
\usepackage[T1]{fontenc}
\usepackage[utf8]{inputenc}
\usepackage{microtype}
\usepackage{inconsolata}
\usepackage{graphicx}

\usepackage{xcolor}
\usepackage{booktabs}
\usepackage{tcolorbox}
\usepackage{tikz}

\newcommand{\framework}{\textsc{SWE-WhatIf}}

\title{SWE-WhatIf: Simulating Real Users of Coding Agents\\from Recorded Sessions}

\author{Tae Soo Kim$^{1,2}$ \quad Tongshuang Wu$^{3}$ \quad Juho Kim$^{1,2}$ \\
$^1$KAIST \quad $^2$SkillBench \quad $^3$Carnegie Mellon University \\
\texttt{taesoo.kim@kaist.ac.kr}}

\begin{document}

\maketitle

\begin{abstract}
Recorded sessions between users and coding agents offer a way to evaluate how agents complete real-world tasks and interact with users. However, these sessions cannot simply be replayed with new agents, because recorded user replies may no longer apply when an agent takes a different trajectory. We introduce \framework{}, which replays recorded sessions with new agents by modeling the user's intents, the criteria for satisfying them, and the conditions under which they arise. Using this model, a user simulator generates replies that adapt to unseen agent turns while remaining grounded in the recording. Across 89 real-world sessions, \framework{} generated the replies most similar to those of real users and matched the strongest baseline when sessions were rewritten. To test adaptation, we replay sessions with agents instructed to avoid or repeat actions that the recorded user pushed back on. \framework{} produced the highest win rate for avoiding these actions (66.3\%), significantly above chance. We further demonstrate its utility by ranking five LLMs as coding agents, where the highest-ranked model depended on how users specified and developed their intents, and a study with 16 users, where replaying their own sessions changed 14 participants' rankings and informed their future agent use.
\end{abstract}

\section{Introduction}

\begin{figure}[t]
\centering
\includegraphics[width=\columnwidth]{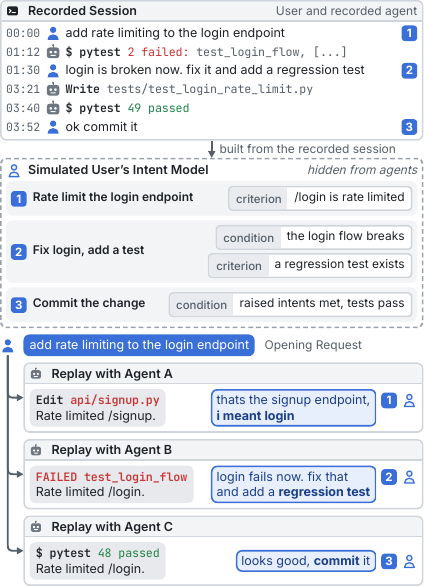}
\caption{Example simulated replays of a recorded session with three agents. \framework{} builds an intent model of the user from the recorded session (top), and the simulated user replies to each agent's turn from this model (bottom).}
\label{fig:teaser}
\end{figure}

Developers increasingly use coding agents in real-world practice.
Recent datasets release recordings of these sessions (e.g., SWE-chat~\citep{baumann2026swe}, DataClaw~\citep{dataclaw2026dataclaw}) that capture both the work users ask agents to do and how they steer them: what outcomes or information they request, what work they accept or reject, and what leads them to change direction or give up.
If we could \emph{replay} these sessions with new agents, agents could be evaluated on their capability to complete real-world tasks while following real users' steering, and users could test which agent best suits their own work.

However, to replay a session, simply resending the user's recorded messages to an agent is not enough.
Each message responds to what the recorded agent did, like a user correcting an agent for prematurely committing a code change.
If a new agent diverges from this observed trajectory (e.g., does not commit the code early), the real user would have replied differently, and the recorded message would no longer match the conversation.
Prior user simulators can respond to unseen agent trajectories~\citep{yao2025tau,wu2025collabllm,vijayvargiya2026ambig,laban2026llms,raghavendra2026swe,trinh2026hilbench}, but typically follow synthetic goals or personas that do not capture a particular user's work or steering.
SWE-Together~\citep{wu2026swe} instead derives a simulator from a conditional script of the user's recorded messages, which can choose when to send each message to a new agent but is constrained to their recorded content and order.
Thus, a replay needs a simulator that can \emph{adapt} to trajectories that the recorded agent never took while remaining \emph{faithful} to why and how the real user steered it.

We introduce \framework{}, a framework that turns a recorded session into a replay task for new coding agents.
Instead of a script of the recorded replies, \framework{} extracts an \emph{intent model}: the outcomes or information the user wanted (\emph{intents}), the \emph{criteria} by which they accepted or rejected the agent's work, and the \emph{conditions} under which each intent was raised or dropped.
During replay, the simulator tracks these intents and, after each agent turn, judges the agent's actions against their criteria and conditions, updates the intent state, and generates the user's next reply based on this state.
As a result, the simulator reacts to what the new agent actually did based on reasons grounded in the recording (e.g., objecting to a commit only if it is actually early).

We evaluate \framework{} both intrinsically, by testing its faithful adaptability against three other user simulators, and extrinsically, by demonstrating its utility in two applications.
For the intrinsic evaluation, we use 89 real-world sessions released by SWE-Together~\citep{wu2026swe} and SWE-chat~\citep{baumann2026swe}.
For replication and robustness, we compare simulated replies with real users' replies on the original recordings and on versions rewritten in form but not in content.
\framework{} generated the most similar replies on the recordings, while matching the best baseline on the rewritten sessions.
For adaptation, we replay each session with an agent instructed to avoid actions that the real user pushed back on and another instructed to repeat them.
With \framework{}, the former agent achieved the highest win rate among the simulators in terms of task success and user burden, significantly above chance.

For the extrinsic evaluation, we first use \framework{} to replay 58 tasks with five LLMs as coding agents in the same harness~\citep{anomaly2025opencode}.
Our ranking agreed more closely with Agent Arena's ranking from real-user sessions than with the other published rankings.
The rankings also differed based on how users steered the agents (e.g., giving details upfront or raising requests only in certain situations).
We then study whether replaying developers' own sessions can help them decide which coding agents to use and how to use them.
In a user study, 16 coding-agent users provided their own sessions and reviewed simulated replays with three different agents.
After reading the replays, 14 of 16 participants changed how they initially ranked the agents, including 10 who changed their top-ranked agent.
From the replays, participants also reported gaining insights for their future agent use---e.g., when cheaper models were sufficient, how each agent works differently, and how to steer agents.

Our contributions are as follows:
\begin{itemize}
    \item We introduce \textbf{\framework{}}, which turns recorded sessions into user simulators that faithfully adapt to new agent trajectories.
    \item We propose \textbf{evaluation methods} to evaluate the faithful adaptability of user simulators by building tests from recorded sessions.
    \item We demonstrate the framework's \textbf{utility} by showing that LLM rankings differ based on user steering patterns and that replaying one's own sessions can help real users choose agents for their own tasks.
\end{itemize}

\section{\framework{}}
\label{sec:system}

\subsection{Task Definition}
\label{sec:replay}

\framework{} creates a \emph{replay task} from a recorded session (i.e., user messages, agent messages, and tool calls) and its \emph{starting environment} (i.e., code repository and runtime when the session began).
In this task, a new agent is given the starting environment and the user's opening request, and it must complete the user's work from the recorded session while interacting turn by turn with a user simulator that steers it as the original user would.
As the new agent can take any path, the simulator must respond to actions absent from the recording, but only in ways supported by the user's recorded requests and reactions.
We refer to this as \emph{faithful adaptability}.

\subsection{User Simulator}
\label{sec:user-simulator}

Inspired by prior work on user simulators driven by representations of user intents~\citep{schatzmann2007agenda,kim2026discoverllm,tack2026llms}, \framework{} builds an \emph{intent model} from the recording to represent the outcomes the user sought and the conditions under which they raised or dropped requests.
During replay, the simulator uses this model to decide which requests and reactions apply to the new agent's actions.

\paragraph{Intent Model.}
The intent model describes the user's work as outcomes rather than turns.
The model consists of a set of \emph{intents}, each an outcome or piece of information that the user wanted during the session (e.g., a bug fixed, an explanation of why a test fails).
Each intent has \emph{criteria} that define what is needed to satisfy it (e.g., the fix must not skip the failing test).
Each intent can also have \emph{conditions} that capture what led the user to raise it: a dependency on another intent (e.g., creating a test after implementing a feature), a situation that arose in the session (e.g., a test failing), or information that the user gained from the agent (e.g., the agent's assessment of a proposed fix).
Conditions can also capture when the user \emph{drops} an intent (e.g., a new fact leads them to give up one approach for another).
Every criterion and condition is grounded in the user's own turns or in outcomes that they implicitly accepted.

\paragraph{Intent State.}
During a replay, the user simulator tracks and updates the state of the intents in the model at each turn.
Each intent is \emph{blocked} (i.e., its conditions are not yet met), \emph{open} (i.e., the simulated user can now express it), \emph{raised} (i.e., the simulated user has expressed it), \emph{dropped} (i.e., it no longer applies), or \emph{abandoned} (i.e., the agent has stopped making progress on it, so the simulated user gives up).
Separately, each criterion of an intent is either \emph{met} or \emph{unmet}, so an intent can be fully, partly, or not yet satisfied.
At the start of a replay, the intents in the user's opening request are already raised.

\paragraph{Style Profile.}
To generate each reply as the user would, the simulator also uses a \emph{style profile} computed from the user's messages in the recording (e.g., length, punctuation, formality).

\paragraph{Building the Intent Model.}
\framework{} builds the intent model with LLM-based agents that draft, review, and revise it.
A drafting agent reads all user messages in the recorded session and groups them into intents with their criteria and conditions.
A reviewing agent then checks the draft for issues, such as overly prescriptive criteria that would penalize functionally equivalent approaches (e.g., requiring a particular function name) or criteria and conditions that cannot be satisfied in a replay (e.g., requiring an external private API).
A final agent revises the draft based on this review.
The style profile is computed directly from the user's messages.
Further details are in Appendix~\ref{app:construction}, and an example intent model is in Appendix~\ref{app:example}.

\subsection{Turn-by-Turn Simulation}
\label{sec:turn}

\begin{figure*}[t]
\centering
\includegraphics[width=0.95\textwidth]{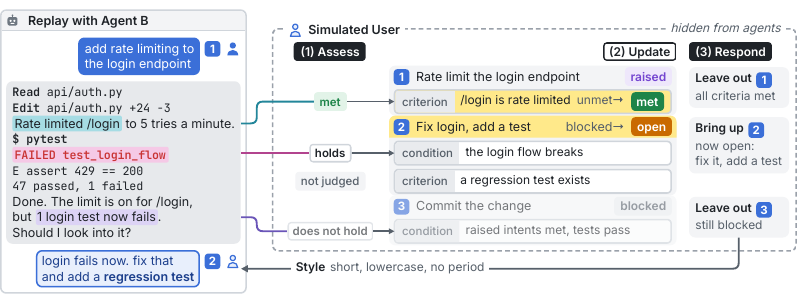}
\caption{After Agent B's turn in Figure~\ref{fig:teaser}, the simulator responds by (1) assessing the criteria of the raised intents and the conditions of blocked intents against the agent's turn, (2) updating the intent states accordingly, and (3) responding by deciding which intents to express in the next message and writing this message in the user's style.}
\label{fig:user-simulator}
\end{figure*}

After each agent turn, our user simulator takes three steps to generate a message (Fig.~\ref{fig:user-simulator}).
\textbf{(1) Assess:} Inspired by LLM-as-a-Judge approaches~\citep{zheng2023judging,ye2024flask,kim2024evallm, viswanathan2025checklists}, we use an LLM to judge the agent's actions and message against the criteria of the raised intents and the conditions of the blocked intents.
\textbf{(2) Update:} Then, based on the judgments, the simulator updates each intent's state (e.g., an intent becomes open when its conditions are met, or abandoned if its criteria do not change for several turns).
\textbf{(3) Respond:} An LLM selects which intents to express next and writes the user's reply, guided by the style profile.
The replay ends when every intent is satisfied, dropped, or abandoned, or when no intent changes state for several consecutive turns (details in Appendix~\ref{app:simulation}).

\subsection{Scoring a Replay}
\label{sec:scoring}

We score each replay on two aspects: \emph{success} (i.e., whether the new agent completed the user's work) and \emph{user burden}, since two replays can reach the same outcome but one can require more steering from the user~\citep{shen2026scaling}.
To judge success, we adapt the goal-based judging used in SWE-Together~\citep{wu2026swe}: an LLM writes a list of \emph{goals} for each task from the recorded session alone, and an LLM judge marks each goal as met or not met from the replay's final state (i.e., repository changes and the agent's messages).
We calculate the percentage of goals met as success, $s$.
To account for burden, we multiply $s$ by $(n_{\mathrm{real}} + 1) / (n_{\mathrm{sim}} + 1)$, where $n_{\mathrm{sim}}$ and $n_{\mathrm{real}}$ are the numbers of messages that the simulated user and the real user sent after the opening request.
This \emph{burden-adjusted success} penalizes a replay that needed more messages than the real session and rewards one that needed fewer.

\section{User Simulator Validation}
\label{sec:validation}

To validate our user simulator, we evaluate three dimensions of \textit{faithful adaptability}.
First, the simulator should reproduce the real user's behavior in the recorded sessions (\textbf{replication}).
Second, it should keep this behavior when a session changes in form but not in content (\textbf{robustness}), such as when turns are paraphrased.
Finally, when the agent's trajectory changes in content, the simulator should respond differently but only as the user would (\textbf{adaptation}).

To evaluate these dimensions, we build tests from the recorded sessions.
For replication, we follow prior work~\citep{dou2025simulatorarena,wu2026humanlm} and compare each simulator's reply with the real user's reply at each point of a session.
For robustness, we repeat this comparison on versions of the sessions with turns paraphrased and merged.
For adaptation, the recorded session does not show how the user would reply to a different agent, and an agent can depart from the recording in too many ways to test each one.
However, the session reveals agent actions that the user pushed back on (e.g., a change that was later undone), and if the agent had avoided such an action, the user would not have needed to push back.
Based on this, we replay each session with an agent that repeats these actions and one that proactively avoids them, where a faithful simulator should produce replays with the latter agent that achieve higher burden-adjusted success.

These evaluations test \framework{} \emph{intrinsically}.
We also conduct \emph{extrinsic evaluations} to demonstrate the framework's utility: ranking LLMs as coding agents and examining how their rankings vary across sessions (\S\ref{sec:agents}), and studying whether replays of their own sessions can help users decide which agents to use and how to use them (\S\ref{sec:user-study}).

\subsection{Setup}

Details on the setup, method, results, and prompts are in Appendix~\ref{app:experiments} and~\ref{app:prompts}.

\paragraph{Tasks.}
We build tasks from recorded sessions that meet three requirements.
First, the recording must show what the user wanted and how they reacted to the agent, so that an intent model can be built from it (e.g., contains real user messages and results of agent tool calls).
Second, a new agent must be able to redo the work from the starting environment alone (e.g., without external resources like GitHub pages or APIs).
Third, the user must have pushed back on at least one agent action, as the adaptation test builds on these actions.
Based on these requirements, we use 65 of the 109 sessions released by SWE-Together~\citep{wu2026swe} and 24 sessions from a newer release of SWE-chat~\citep{baumann2026swe}, for 89 tasks in all.

\paragraph{User Simulator Designs.}
We compare \framework{}, with and without the style profile, against three other user simulators, all run on the same LLM (GPT-5.6 Luna).
For \textbf{SWE-Together}~\citep{wu2026swe} tasks, we use the simulator prompt released for each of its tasks, which lists the user's later messages and in what conditions each one should be sent.
For the SWE-chat tasks, we use an LLM (Claude Opus 5.5) to write prompts in the same form.
This simulator also keeps a summary of each earlier agent turn.
\textbf{SWE-Interact}~\citep{raghavendra2026swe} role-plays a terse colleague who reveals a list of requirements one at a time.
SWE-Interact derives this list from benchmark tasks (e.g., SWE-bench~\citep{jimenez2024swe}).
As our tasks come from recorded sessions, we use the intents and criteria from our intent model instead.
Finally, the \textbf{goal-prompted simulator} receives the same intents and criteria, but without a persona.

\subsection{Method}
\label{sec:method}

Each test compares simulators on the same tasks.

\paragraph{Replication.}
At every user reply in the recorded sessions, we give each simulator the conversation up to that reply and have it generate the next reply.
Simulators that keep a state across turns also receive the state built from the conversation so far (i.e., the intent states of \framework{} and summaries of earlier turns for SWE-Together).
An LLM judge (GPT-6.1 Sol) rates the \textbf{similarity} of each generated reply with the real user's reply on content and style from 1 to 5, adapting the similarity ratings of SimulatorArena~\citep{dou2025simulatorarena}, and we scale the mean of the two ratings to 0--100.
In a replay, however, each reply builds on the simulator's own earlier replies rather than the real user's.
To test this, we also use a \textbf{multi-turn} setting, where each simulator writes three consecutive replies between the recorded agent turns, and the judge rates the three against the real user's three replies.

\paragraph{Robustness.}
We rewrite each recorded session by using an LLM to paraphrase all turns and merging some consecutive user turns.
We then run the same single-turn and multi-turn tests on the rewritten sessions, to check that a simulator does not rely on the exact wording and order of the recorded turns.

\paragraph{Adaptation.}
For each task and user simulator, we replay the session with two agents (Claude Code on Claude Sonnet 5.5): one instructed to repeat the actions that the user pushed back on (\emph{observed policy}) and one instructed to avoid them (\emph{proactive policy}).
Following prior work~\citep{baumann2026swe,tang2026how,tang2026programming}, we use an LLM (GPT-5.6 Terra) to identify these actions (e.g., ones that the user corrected, objected to, or reverted) and to write the instructions for each policy.
For each user simulator, we replay sessions with each agent policy twice.
We compare the burden-adjusted success (\S\ref{sec:scoring}) of each proactive-policy replay with each observed-policy replay of the same task, and report the \emph{win rate} of the proactive policy on a scale of 0 to 100, counting a tie as half a win~\citep{vargha2000critique}.
We use two-sided sign tests to assess whether the proactive policy wins significantly more often than it loses, with Holm correction across the five simulators~\citep{holm1979simple}.

\begin{table*}[t]
\centering
\small
\begin{tabular}{lccccccc}
\toprule
& \multicolumn{3}{c}{\textbf{Replication}} & \multicolumn{3}{c}{\textbf{Robustness}} & \textbf{Adaptation} \\
\cmidrule(lr){2-4} \cmidrule(lr){5-7} \cmidrule(lr){8-8}
\textbf{User Simulator} & \textbf{Single} & \textbf{Multi} & \textbf{Avg.} & \textbf{Single} & \textbf{Multi} & \textbf{Avg.} & \textbf{Win Rate} \\
\midrule
Goal-Prompted & 26.3 & 24.8 & 25.6 & 25.9 & 24.4 & 25.2 & 55.1 ($p$=.306) \\
SWE-Interact & 35.0 & 31.4 & 33.2 & 35.8 & 30.3 & 33.1 & 50.6 ($p$=1.00) \\
SWE-Together & 51.3 & 40.4 & 45.9 & \underline{48.1} & \textbf{41.3} & \underline{44.7} & 55.6 ($p$=.314) \\
\midrule
\textbf{\framework{}} & \textbf{53.2} & \textbf{44.9} & \textbf{49.1} & \textbf{49.5} & \underline{40.7} & \textbf{45.1} & \textbf{66.3} ($p$=.001) \\
\quad $-$ Style profile & \underline{51.8} & \underline{42.0} & \underline{46.9} & 47.3 & 40.0 & 43.7 & \underline{62.9} ($p$=.012) \\
\bottomrule
\end{tabular}
\caption{Results of each user simulator on replication and robustness (similarity to the real user's replies for single replies, three consecutive replies, and their average) and on adaptation (win rate of the proactive policy), all scaled to 0--100. $p$ is from a two-sided sign test before Holm correction. Best results are \textbf{bold}-faced and second best results are \underline{underlined}.}
\label{tab:results}
\end{table*}

\begin{figure}[t]
\centering
\includegraphics[width=0.80\columnwidth]{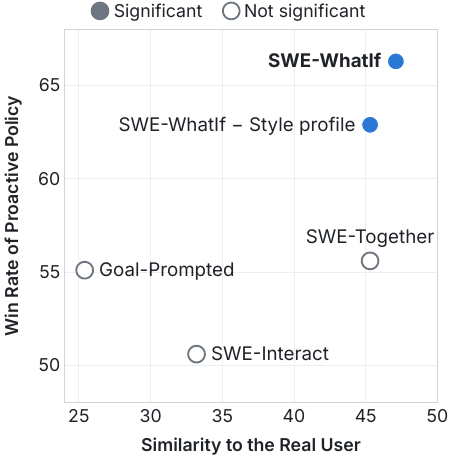}
\caption{Replication and robustness score averages (i.e., similarity) against adaptation score (i.e., win rate of the proactive policy), per user simulator. Filled points are simulators with significant results for adaptation.}
\label{fig:faithful-adaptability}
\end{figure}

\subsection{Results}
\label{sec:results}

Table~\ref{tab:results} gives the results of the three tests, and Figure~\ref{fig:faithful-adaptability} plots each user simulator's scores on the adaptation test against the average of its replication and robustness scores.
\framework{} scored at least as high as SWE-Together on replication and robustness, while also being the simulator with which the proactive policy reached the highest win rate and won significantly more tasks than it lost.
Thus, \framework{} showed the most faithful adaptability among the simulators, replying as the real user did on the recorded sessions while still responding to a different agent as the user would.

\paragraph{Replication.}
\framework{} scored highest on the recorded sessions, both for single replies (53.2, against 51.3 for SWE-Together) and for three consecutive replies (44.9 against 40.4), even though SWE-Together's prompt provides the recorded user's later messages verbatim.
In comparison, SWE-Interact and the goal-prompted simulator scored lower in both settings.
All simulators scored lower in the multi-turn setting than in single-turn.

\paragraph{Robustness.}
On the rewritten sessions, \framework{} and SWE-Together scored about the same (45.1 and 44.7 on average).
Both scored lower than on the recorded sessions, while SWE-Interact and the goal-prompted simulator were more robust to the paraphrased and merged turns, which can be expected as their replies are already more general and less tied to the content of the recording.

\paragraph{Adaptation.}
\framework{} led to the highest win rate for the proactive policy (66.3), and removing the style profile lowered it to 62.9 (Table~\ref{tab:results}).
With the two versions of \framework{}, the proactive policy won 52 tasks and lost 23 with the style profile, and won 50 and lost 27 without.

\paragraph{Error Analysis.}
To understand when and why the user simulators did not distinguish the two policies in the adaptation test, we first inspected the simulated user turns in tasks where the replays tied or favored the observed policy, and identified four kinds of error in the simulator.
We then used an LLM (Claude Sonnet 5.5) to label the first replay with each policy in every task (Table~\ref{tab:errors}).
As expected, the goal-prompted simulator tended to state all of its intents in its first turns, so the replays with both policies were similarly successful and took few turns.
SWE-Interact and \framework{} more often asked for extra work (e.g., pasting a diff, reviewing the changes, or running the checks again), which the real user never requested in the recorded session.
When the replayed agent departed from the recorded trajectory, SWE-Together had no scripted message that fit, so it mostly replied passively by accepting what the agent did or telling the agent to continue without steering the session.

\begin{table}[t]
\centering
\footnotesize
\setlength{\tabcolsep}{3.5pt}
\begin{tabular}{lcccc}
\toprule
\textbf{User Simulator} & \textbf{Premature} & \textbf{Extra} & \textbf{Passive} & \textbf{Omitted} \\
\midrule
Goal-Prompted & \textbf{77.5\%} & 64.0\% & 1.69\% & 18.5\% \\
SWE-Interact & 53.4\% & \textbf{70.2\%} & 3.37\% & 23.6\% \\
SWE-Together & 12.9\% & 44.9\% & \textbf{96.1\%} & 35.4\% \\
\midrule
\textbf{\framework{}} & 33.1\% & \textbf{52.2\%} & 5.62\% & 21.3\% \\
\quad $-$ Style profile & 31.5\% & \textbf{47.2\%} & 4.49\% & 25.3\% \\
\bottomrule
\end{tabular}
\caption{Share of each user simulator's replays with at least one turn showing each kind of error (89 tasks, the first replay with each policy, 178 replays per simulator). A \emph{premature} turn asks for something before the user did in the recording, an \emph{extra} turn asks for work that the real user never asked for, a \emph{passive} turn lets the agent continue where the real user stepped in, and an \emph{omitted} turn leaves out a request of the real user that still applies. Each simulator's most common error is \textbf{bold}-faced.}
\label{tab:errors}
\end{table}

\section{Evaluating LLMs as Coding Agents}
\label{sec:agents}

We use \framework{} to rank LLMs as coding agents by how well they serve the users in the recorded sessions.
We replay 58 of the 65 tasks from SWE-Together's sessions once with five LLMs (Claude Opus 5.5, Claude Sonnet 5.5, GPT-6 Sol, GPT-6 Luna, and Kimi K3) in the same agent harness, opencode~\citep{anomaly2025opencode}---excluding seven tasks where an agent could not finish within the time limit.
We then rank the models by their mean burden-adjusted success over the tasks (\S\ref{sec:scoring}).
Details are in Appendix~\ref{app:agents}.

\begin{figure*}[t]
\centering
\includegraphics[width=0.9\textwidth]{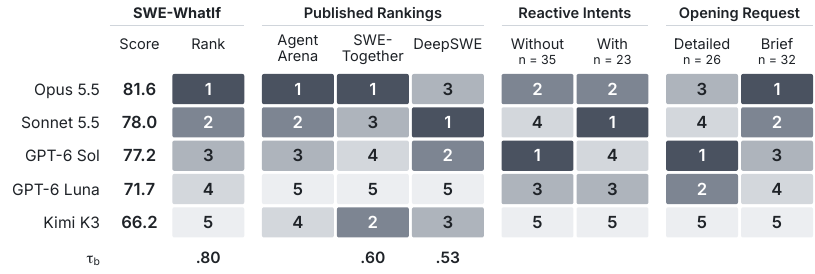}
\caption{Mean burden-adjusted success and rank of each model with \framework{} over the 58 tasks (Score), and the rank of each model under three published rankings, and within each group of tasks, with darker cells for better ranks. Tied models share a rank, and $n$ gives the number of tasks in each group. The last row gives the agreement of each ranking with Agent Arena's (Kendall's $\tau_b$).}
\label{fig:agent-ranks}
\end{figure*}

\paragraph{Comparison with Existing Rankings.}
We compare our ranking with three published rankings that differ in user involvement: DeepSWE~\citep{huang2026deepswe} (i.e., no user), SWE-Together's leaderboard~\citep{wu2026swe} (i.e., simulated users), and Agent Arena~\citep{chiang2024chatbot,arena2026agent} (i.e., real users' sessions with coding agents).
Our ranking agreed with Agent Arena's more closely ($\tau_b$=.80) than SWE-Together's ($\tau_b$=.60) and DeepSWE's ($\tau_b$=.53) did (Figure~\ref{fig:agent-ranks}), which suggests that it could more closely reflect how real users judge coding agents.
As seen in Appendix~\ref{app:agents}, Claude Opus 5.5 and Claude Sonnet 5.5 met about as many goals (65.3\%, 65.1\%), but the user required fewer messages with Opus (6.4 and 6.7 on average).

\paragraph{Rankings across User Groups.}
To demonstrate how the best agent can differ between users or sessions, we group the tasks by two properties of their intent models and rank the agents within each group (Figure~\ref{fig:agent-ranks}).
First, we group the tasks by whether they involve \emph{reactive intents} (i.e., intents that the user raises only if a specific situation arises during the replay, like a test or a review that suggests another change).
We also split the tasks by whether their opening request is \emph{detailed} (i.e., states more than two criteria of its intents; 26 tasks), or \emph{brief} (32 tasks).
GPT-6 Sol ranked first on tasks without reactive intents and fourth on tasks with them.
This suggests that the Claude models better handle tasks where additional intents become relevant during the agent's work, while GPT-6 Sol performs better when no intents depend on situations that arise during the replay.
With detailed requests, the models met similar numbers of goals and the GPT models needed fewer messages, while with brief requests the Claude models met more goals.
This suggests that the GPT models more efficiently complete tasks whose requirements are stated upfront, while the Claude models perform better when requirements are raised later.

\section{User Study}
\label{sec:user-study}

\framework{} can allow users of coding agents to replay their own sessions with different agent harnesses and models, which could show them which agents best handle their tasks and steering patterns.
To understand what users learn from such replays, we conducted a study with 16 users of coding agents.

\paragraph{Method.}
We recruited participants through Prolific.
In our study interface, each participant provided one of their own recorded sessions, from which a \framework{} user simulator was built and used to replay part of the session with three agents for up to three turns each.
Before seeing the replays, participants first predicted how each agent would rank in the replays.
Then, they walked through the replays side by side where the agent used in each replay was hidden, and they then ranked the anonymous agents.
At the end, each replay's agent was revealed and participants were asked what they learned from the replays and whether this would change how they use agents in the future.
Participants also rated how closely the simulated user's replies represented how they steer agents and write messages.
Details in Appendix~\ref{app:user-study}.

\begin{figure}[t]
\centering
\includegraphics[width=\columnwidth]{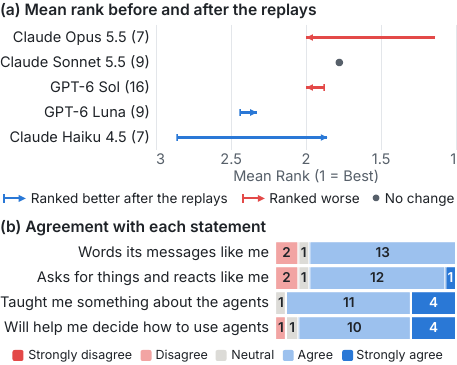}
\caption{(a) Mean rank that participants gave each agent before reading the replays (start of each arrow) and after (arrowhead), where 1 is best. The number of participants who saw each agent is in parentheses. (b) Participants' agreement with the four statements about their user simulators and the replays (16 participants).}
\label{fig:study-summary}
\end{figure}

\paragraph{Results.}
After reading the replays, 14 of 16 participants ranked the agents differently from their initial prediction, and 10 changed which agent they ranked first.
On average, participants ranked the smaller model of each set higher than they had predicted, and Claude Opus 5.5 and GPT-6 Sol lower (Fig.~\ref{fig:study-summary}a).
Fifteen agreed that the replays taught them something they did not know about the agents, and 14 agreed that the replays would help them decide how to use coding agents (Fig.~\ref{fig:study-summary}b).

The most common lesson was that a smaller and cheaper model was enough for their session (12 of 16)---e.g., ``Luna performed as well as Sol at a fraction of the cost.''
Participants also saw that agents differed in how they worked with the user, and this could affect which agent they preferred (8 of 16).
For example, P04 preferred an agent that ``listened to what I had to say'' over one that kept proposing its own ideas.
Also, 4 participants confirmed or revised their harness preferences (e.g., Claude Code or Codex), and 2 learned that their steering shapes the result.
For future use, 11 of 16 planned to choose models by task, including 5 who would start with a smaller model, 4 planned to change their steering, and 3 would keep their current harness.

Thirteen of 16 participants agreed that their simulated selves worded messages like them, and 13 agreed that they asked for things and reacted to the agent like them.
Participants marked 49\% of the simulated replies as ``Sounds like me'', 38\% as ``Partly'', and 13\% as ``Not like me''.
Where the simulated user differed, 5 participants found it more formal or wordier than themselves (e.g., P15: ``with a lot more formality than I would''), and 3 found that it kept insisting on the same thing when they would not have (Appendix~\ref{app:study}).

\section{Related Work}
\label{sec:related}

\paragraph{Benchmarks for Coding Agents.}
Benchmarks for coding agents mainly assess fully specified tasks without a user~\citep{wang2026humans}, such as SWE-bench~\citep{jimenez2024swe} and AppWorld~\citep{trivedi2024appworld}.
As coding agents are increasingly used in practice, recent work adds simulated users to these benchmarks~\citep{pan2025benchmarks,kim2025codeassistbench,trinh2026hilbench,wang2024mint} who reveal requirements when asked~\citep{vijayvargiya2026ambig} or one at a time~\citep{raghavendra2026swe}.

\paragraph{User Simulators.}
User simulation in task-oriented dialogue~\citep{schatzmann2006survey} includes agenda-based approaches~\citep{schatzmann2007agenda}.
Recent work prompts LLMs with goals~\citep{yao2025tau,barres2026tau,wu2025collabllm} or personas~\citep{raghavendra2026swe}, or trains them to simulate users~\citep{zhou2026odyssim,sun2026reinforcing,phan2026mimesis,wang2026learning}.
Further, simulators have also been designed to model specific user behaviors, such as progressive disclosure~\citep{laban2026llms}, evolving intents~\citep{kim2026discoverllm,tack2026llms}, contextual preferences~\citep{kim2025cupid}, knowledge~\citep{lee2026knowsim}, and cooperation~\citep{chopra2026beyond}.
Others build simulators from real user data~\citep{wu2026humanlm,naous2026flipping}, including inferred profiles~\citep{wang2025know,zhou2026tom}, but simulated users can still differ from real users in their interaction behavior and failure patterns~\citep{zhou2026mind,seshadri2026lost}.

\paragraph{Evaluation from Recorded Sessions.}
Recorded coding-agent sessions~\citep{baumann2026swe,dataclaw2026dataclaw,he2026recap} support analyses of user steering~\citep{tang2026how,tang2026programming} and evaluation on real tasks.
SWE-Together~\citep{wu2026swe} replays these sessions by deciding when to send the recorded user's messages, while preserving their content and order.
Instead, \framework{} generates replies from the user's intents, criteria, and conditions, allowing it to respond to new agent trajectories.

\section{Conclusion}

We introduce \framework{}, a framework for replaying recorded user--agent sessions with new coding agents by simulating users from their intents, criteria, and conditions.
In tests of replication and robustness, \framework{} generated the replies most similar to those of real users on recorded sessions and matched the strongest baseline when sessions were rewritten.
In an adaptation test, our simulator most clearly distinguished agents that avoided actions that the real users had objected to from those that repeated them.
By ranking five LLMs as coding agents and through a user study where participants replayed their own sessions, we demonstrate that replays can help researchers evaluate agents on real users' work and help developers choose agents for their own tasks.

\section*{Limitations}
Our replays make the following assumptions about users.
First, the user interacts with the agent only through the conversation.
In practice, users may also open and edit files, run tests on their own, or look at external resources, but these are not shown in the recorded session.
Second, the user implicitly accepted any outcome they did not object to, which ignores cases where the user might have been dissatisfied but did not express it.
Third, the simulated user does not react to anything for which the recorded session has no relevant evidence (e.g., if a new agent adds comments to the code but the user never reacted to code comments in the recorded session, the simulated user ignores them).
Similarly, the simulated user only asks for what the user asked for in the recording, even when a new agent's path might have led the real user to new requests.
Finally, the simulated user reads every tool call and its result, while a real user may only skim the agent's work.
We leave relaxing these assumptions for future work.

Our experiments also have the following limitations.
The used sessions come from public releases, so they include only developers who chose to publish their sessions (e.g., SWE-chat~\citep{baumann2026swe} mostly includes early adopters working on open-source code, not on proprietary or enterprise code).
For agent evaluation and ranking, we evaluated only five models in a single agent harness, opencode~\citep{anomaly2025opencode}.
Our metric for burden-adjusted success approximates the user's effort of writing and reading by counting messages, and the goals used to measure success are based on where the recorded session ended, but the user may have stopped midway instead of reaching the state they wanted.
In the user study, participants saw only three agents that we chose, for up to three turns each, and these agents worked in environments partially reconstructed from the recording, which may not reflect their full capabilities.

\section*{Ethical Considerations}

Our experiments use recorded sessions released publicly by SWE-Together \citep{wu2026swe} and SWE-chat \citep{baumann2026swe}.
SWE-Together is released under the Apache 2.0 license, with sessions originally collected from SWE-chat (ODC-By), DataClaw (MIT), Pi, and Hyperswitch.
We use these sessions to evaluate coding agents and user simulators, consistent with the purposes of these releases.
However, recorded sessions can contain sensitive information, such as private code, credentials, and identifying details.
Building user simulators from these sessions can further expose this information through their intent models, style profiles, and simulated replays.

In our user study, participants gave informed consent before providing their sessions.
Before uploading, we replaced credentials and identifying details on participants' own computers and sent only the portions of sessions needed for the replays.
Participants were informed that their sessions would be sent to the model providers that ran the agents and user simulator.
After generating the replays, we deleted the uploaded sessions, intent models, style profiles, and starting environments.
We do not release an artifact package containing the source sessions, code, intent models, style profiles, goals, or simulated replays.
The study was approved by KAIST's institutional review board.

Beyond privacy, \framework{} raises another concern because it simulates how a particular user communicates with and steers coding agents.
Although we use these simulators only to evaluate coding agents, they could also be used to imitate the real users in other settings without their knowledge or consent.
This risk is especially relevant when sharing intent models, style profiles, or simulated replays that capture how a particular user behaves.

We used AI assistants to help implement the user simulator, build the user study interface, revise the manuscript, and search the literature.
The authors verified and revised all resulting content and read the cited papers.

\bibliography{references}

\clearpage{}

\appendix
\renewcommand{\topfraction}{0.9}
\renewcommand{\dbltopfraction}{0.9}
\renewcommand{\floatpagefraction}{0.8}
\renewcommand{\dblfloatpagefraction}{0.8}
\renewcommand{\textfraction}{0.1}

\section{User Simulator Details}
\label{app:simulator}

Here, we provide the details of how \framework{} builds the intent model of each task and how the user simulator replies after each agent turn.

\subsection{Building the Intent Model}
\label{app:construction}

We build each intent model with three LLM-based agents that draft, review, and revise it in turn, all Claude Code on Claude Opus 5.5.

\paragraph{Input.}
Each agent takes the recorded session (i.e., every message of the user and the agent, and the results of the agent's tool calls) and also the starting environment (i.e., the code repository with its history and its runtime).

\paragraph{Drafting.}
The drafting agent writes the intent model through five steps (prompt in Fig.~\ref{fig:prompt-draft}, with the rules in Fig.~\ref{fig:prompt-rules}): (1) label what each user message does (e.g., a request, a correction, or an acceptance of the agent's work); (2) group messages into intents and write the criteria of each intent; (3) for each criterion, record the evidence in the session and what the simulated user could observe in a replay to judge it; (4) write the conditions; and (5) mark where in each session each criterion and condition was met.

\paragraph{Main Rules.}
We instruct the drafting agent to follow specific rules that tie the intent model to the user's own requests and decisions.
The following are the main rules that are followed:
\begin{itemize}
    \item An intent or criterion should refer to names that were only named explicitly by the user, given in the opening request, or already held by the starting environment.
    \item A criterion can require a specific approach to code changes only if the user requested that approach explicitly or they explicitly accepted when an agent took that approach. Otherwise, the criterion must be described so that it can be satisfied by other equivalent approaches.
    \item A condition records what led the user to raise an intent in the conversation.
    \item A question and the request that follows its answer are two intents, where the second has a condition that depends on the first.
    \item If new information leads the user to replace one request with another, represent the original and replacement requests as separate intents.
    Give the original intent a condition that drops it when that information appears in the replay.
    If the user instead refines what they want from the same outcome, without learning new information, keep one intent and write its criteria to reflect the user's final requirements.
\end{itemize}

\paragraph{Review and Revise.}
The reviewing agent reads the session, the draft, and the starting environment, and lists issues without changing the draft (Fig.~\ref{fig:prompt-review}).
While reviewing, it can also run commands in the starting environment.
Finally, a revising agent fixes each listed issue (Fig.~\ref{fig:prompt-revise}).

\paragraph{Statistics.}
Table~\ref{tab:intent-stats} shows statistics of the 89 recorded sessions used in the experiments.
Of the 491 intents across all sessions, 291 had a condition for raising them and 6 had a condition for dropping them.
Most conditions (137 of 230) only require another intent to be satisfied.
Sessions with more user messages have larger intent models, with Spearman's $\rho$ of 0.79 for the number of intents, 0.64 for criteria, and 0.71 for conditions.
However, the number of characters per message does not relate to the size of the intent model ($|\rho| \leq 0.17$).

\begin{table}[t]
\centering
\small
\begin{tabular}{lrrr}
\toprule
 & \textbf{Median} & \textbf{Min} & \textbf{Max} \\
\midrule
User messages & 6 & 2 & 27 \\
Characters per message & 82 & 21 & 227 \\
\midrule
Intents & 5 & 1 & 16 \\
Criteria & 10 & 2 & 41 \\
Conditions & 2 & 0 & 9 \\
\bottomrule
\end{tabular}
\caption{Median, minimum, and maximum per task of the user's messages in the recorded sessions of the 89 tasks and of their intent models. User messages include the opening request and leave out slash commands (i.e., commands to the agent's interface such as clearing the context). Characters are averaged over the messages that the user typed.}
\label{tab:intent-stats}
\end{table}

\begin{figure*}[t]
\centering
\small
\textbf{(a) Intent model}\par\smallskip
\begin{tabular}{@{}p{0.4cm}p{4.0cm}p{7.9cm}p{2.3cm}@{}}
\toprule
 & Intent & Criteria & Raised when \\
\midrule
\textit{I1} & \texttt{entire explain} prints its help text when given an argument without a qualifier flag. & (a) Print the help text for an argument without any of the qualifier flags (\texttt{-{}-session}, \texttt{-{}-commit}, \texttt{-{}-checkpoint}). (b) Command with no argument, or with a qualifier flag, should behave as it did before the patch. & Opening request \\
\addlinespace
\textit{I2} & The patch is on a new branch. & (a) A new branch is created and holds the changes of the patch. (b) None of the changes is committed outside the new branch. & No condition \\
\addlinespace
\textit{I3} & The code changes are reviewed with a verdict. & (a) The agent reviews the changes of the patch and should decide to approve or request changes. & \textit{I1} satisfied \\
\addlinespace
\textit{I4} & Add a short hint that explains why the help text is shown, if the change is small. & (a) Print a short hint that explains why the help text before it. (b) Adding the hint is a small change that does not affect the rest of the command. & \textit{C1} met \\
\midrule
\multicolumn{4}{@{}p{15.5cm}@{}}{\textit{C1}: A review of the patch suggests printing a message that explains why the help text is shown.} \\
\bottomrule
\end{tabular}

\bigskip
\textbf{(b) Replay excerpt}\par\smallskip
\begin{tabular}{@{}p{0.6cm}p{4.6cm}p{4.0cm}p{2.9cm}p{2.2cm}@{}}
\toprule
Turn & New agent & Assess and update & Simulated user & Intent state after \\
\midrule
1 & Adds a check that prints the help text when an argument is given, adds a test, and runs the tests of the command, which pass. & \textit{I1}: (a) met. (b) unmet, as the check also applies when a qualifier flag is given. \textit{C1} not met. & Asks for the check to apply only without qualifier flags and to keep the behavior with no argument. & \textit{I1} raised with 1 of 2 criteria met, \textit{I2} open, \textit{I3} and \textit{I4} blocked \\
\addlinespace
2 & Limits the check to arguments without a qualifier flag and adds a test for the flag case. & \textit{I1}: (b) met. \textit{I3} becomes open. & Sends \texttt{/reviewer}. & \textit{I1} satisfied, \textit{I3} raised, \textit{I2} open, \textit{I4} blocked \\
\addlinespace
3 & Says that it cannot run \texttt{/reviewer} without a tool to start another agent, and offers to review the changes itself. & No criterion or condition changes. & Asks it to review the changes directly and give a verdict. & Unchanged \\
\bottomrule
\end{tabular}
\caption{Example intent model of one task (a) and the first three agent turns of one of its replays with a new agent (b). All content is paraphrased. Before turn 1, \textit{I1} is raised by the opening request, \textit{I2} is open, and \textit{I3} and \textit{I4} are blocked.}
\label{fig:example-intent-model}
\end{figure*}

\subsection{Simulating a Turn}
\label{app:simulation}

Our user simulator runs on GPT-5.6 Luna at medium reasoning effort. 
The simulator takes as input only the conversation of the replay, which includes the agent's tool calls and their results.

\paragraph{Assess.}
There are two types of conditions: a condition that depends on another intent (e.g., \textit{I1} satisfied) is checked by the simulator from the intent state, and a condition on a situation in the session or on information from the agent is written in text and judged by an LLM.
After each agent turn, an LLM judges the criteria of every raised or dropped intent, and every condition written in text, including those already met (Fig.~\ref{fig:prompt-assess}).
For each judged criterion or condition, the LLM also describes its reasoning for its judgement.

\paragraph{Update.}
The simulator then updates the intent state from the judgments, without an LLM.
At the start of a replay, the intents in the opening request are raised, the intents without a raising condition are open, and all other intents are blocked.
A blocked intent becomes open when its raising conditions are met, and an open intent becomes raised only when the respond step expresses it.
An intent becomes dropped when its dropping conditions are met, and satisfied when all of its criteria are met.
Once a condition is met, it stays met.
In contrast, a met criterion becomes unmet again if a later agent turn does work that undoes or conflicts with the criterion.
A raised intent that is not satisfied is abandoned when none of its criteria change for four consecutive agent turns.
Blocked intents that depend on an abandoned intent are also abandoned.
A dropped or abandoned intent is never raised again.

\paragraph{Respond.}
An LLM then chooses which intents to express among those that are open or already raised but with unmet criteria (Fig.~\ref{fig:prompt-respond}).
The LLM first writes what the reply should ask the agent to do, and then rewrites the reply to follow the user's style with the style profile.
The respond step is also given the assess step's reasoning for the judgement of each unmet criterion, so the simulator can name what is missing (e.g., the agent built the binary but never ran it).

\paragraph{Style Profile.}
We compute the style profile from the messages that the user typed in the recorded session with deterministic code.
These messages include the opening request and leave out slash commands and pasted text (i.e., messages longer than 600 characters or with six or more lines).
The style profile covers the language used, whether the first letter is capitalized, the end punctuation, the use of backticks, the length in words, and the use of specific words (e.g., ``please'', ``pls'', ``thanks'').
For each pattern except length, the code counts the share of messages with the pattern and turns it into a fixed sentence (e.g., ``None of their messages say please or thanks.'').
For length, the code takes the median number of words per message and places it in one of four classes, which are a few words (up to 6), one sentence (up to 15), one to three sentences (up to 35), and several sentences.

\paragraph{Ending a Replay.}
A replay ends when every intent is satisfied, dropped, or abandoned, or when no intent changes for seven consecutive agent turns.
Intents that the user would want only if a situation arose (e.g., a test failing) are left out of this check while they are blocked.

\subsection{Example}
\label{app:example}

Figure~\ref{fig:example-intent-model} shows the intent model of one task and an excerpt of one of its replays.
In the recorded session, the user asks for a small patch to the command-line tool of a Go repository, so that the command \texttt{entire explain} prints its help text when it receives an argument without any of its qualifier flags (i.e., the flags that name what to explain).
When the recorded agent starts to edit the code, the user rejects the edit and asks it to create a branch first.
After the patch, the user runs \texttt{/reviewer} (i.e., a command defined in the repository that has the agent start a separate agent to review the changes).
The review approves the patch and suggests a hint that explains why the help text is shown, and the user asks for the hint if it is small.
The intent model keeps these requests as four intents.
\textit{I2} has no raising condition, so it is open from the start of a replay.
\textit{I3} is raised when \textit{I1} is satisfied, and \textit{I4} is raised only if a review suggests the hint (\textit{C1}).

In the excerpt, the new agent (Claude Code on Claude Sonnet 5.5) makes the patch, but its check also prints the help text when a qualifier flag is given.
As criterion (b) of \textit{I1} is unmet, the simulated user asks for this missing part.
In the third turn, the agent is unable to run \texttt{/reviewer}, as it has no tool to start another agent, which is a situation that was not present in the recording.
Then, the simulated user instead asks it to review the changes itself so the agent provides a review that approves the patch without suggesting a hint.
As a result, \textit{C1} is never met and the simulated user never asks for the hint.
The replay ended after five agent turns, with \textit{I1} to \textit{I3} satisfied.
In another replay with the same agent, the agent also does the same review but it instead suggested including a message that explains the help text.
This led the simulated user to ask for the hint in its next reply.

\section{Experiment Details}
\label{app:experiments}

This appendix gives the details on the validation experiments for the user simulators (Section~\ref{sec:validation}) and of the evaluation of LLMs as coding agents (Section~\ref{sec:agents}).

\subsection{Tasks}
\label{app:tasks}

We build our tasks from the 109 sessions released by SWE-Together~\citep{wu2026swe}, each with a prebuilt starting environment.
SWE-Together's sessions were originally collected in an earlier release of SWE-chat~\citep{baumann2026swe}, DataClaw~\citep{dataclaw2026dataclaw}, Pi~\citep{zechner2026pi}, and Hyperswitch~\citep{archit112025claude}.
The release is under the Apache 2.0 license and states no separate license for its data.
The 89 tasks included the following languages: Chinese in 4 sessions, Vietnamese in 3 sessions, and English in the other 82.

\begin{table*}[t]
\centering
\scriptsize
\setlength{\tabcolsep}{4pt}
\begin{tabular}{lll}
\toprule
\multicolumn{3}{l}{\textbf{SWE-Together (65 tasks, by \texttt{task\_id})}} \\
\midrule
\texttt{agent-swarm-task-4a881b} & \texttt{cli-task-ea3f8f} & \texttt{pi-mono-auto-6f8288bb} \\
\texttt{agent-swarm-task-ea4bd8} & \texttt{comfyui-gemma3-sliding-window} & \texttt{pi-mono-auto-90748a5d} \\
\texttt{amytis-task-e3714e} & \texttt{comfyui-jina-clip-v2-full} & \texttt{pi-mono-auto-93c17d3b} \\
\texttt{arr-monitor-add-processes-flag} & \texttt{comfyui-lumina-axes-lens} & \texttt{pi-mono-auto-95de1efe} \\
\texttt{cli-task-0ec2e9} & \texttt{comfyui-lumina2-lora-prefix} & \texttt{pi-mono-auto-a4fca584} \\
\texttt{cli-task-2a55af} & \texttt{comfyui-newbie-lumina-refactor} & \texttt{pi-mono-auto-cad40af4} \\
\texttt{cli-task-2c3e30} & \texttt{comfyui-triton-windows-amd-fix} & \texttt{pi-mono-auto-cbb62cbe} \\
\texttt{cli-task-2f5833} & \texttt{dataclaw-anonymizer-tests} & \texttt{pi-mono-auto-d3b2130d} \\
\texttt{cli-task-32ba47} & \texttt{dataclaw-windows-path-fix} & \texttt{pi-mono-auto-d8c68afc} \\
\texttt{cli-task-33e050} & \texttt{gemini-voyager-task-16a5c7} & \texttt{pi-mono-auto-ec7037ba} \\
\texttt{cli-task-4a9dde} & \texttt{gemini-voyager-task-aa88f5} & \texttt{pi-mono-extensions-event-refactor} \\
\texttt{cli-task-577e8c} & \texttt{gemini-voyager-task-c5c01d} & \texttt{pi-mono-keybinding-scope} \\
\texttt{cli-task-70c88c} & \texttt{gemini-voyager-task-f519c2} & \texttt{pi-mono-parallel-tool-stall} \\
\texttt{cli-task-726de6} & \texttt{light-protocol-implement-7adad1} & \texttt{pi-mono-tool-execution-write-error} \\
\texttt{cli-task-aa4038} & \texttt{lock-code-manager-fix-7c955a} & \texttt{qwen3-moe-gguf-dequant} \\
\texttt{cli-task-b12319} & \texttt{nunchaku-quantize-bugfix} & \texttt{rudel-task-8e0bd6} \\
\texttt{cli-task-b3d2dd} & \texttt{nunchaku-svdq-reconstruction} & \texttt{sd-scripts-fp8-lumina} \\
\texttt{cli-task-c01017} & \texttt{pi-mono-auto-0467b78e} & \texttt{sd-scripts-multiresolution-compat} \\
\texttt{cli-task-cd4662} & \texttt{pi-mono-auto-0fa6d7bd} & \texttt{sd-scripts-reg-image-dedup} \\
\texttt{cli-task-cf32cb} & \texttt{pi-mono-auto-1b58dd4f} & \texttt{sd-scripts-sdxl-multires-dedup} \\
\texttt{cli-task-dbd2bf} & \texttt{pi-mono-auto-41636ae5} & \texttt{triton-msvc-c4267-warnings} \\
\texttt{cli-task-e5813e} & \texttt{pi-mono-auto-4439324b} &  \\
\midrule
\multicolumn{3}{l}{\textbf{SWE-chat (24 sessions, by \texttt{session\_id})}} \\
\midrule
\texttt{08774980-a9e2-4ef4-a696-0a7e232eafc9} & \texttt{4ddd0676-3c02-4818-a0e3-c13c70fb8d6f} & \texttt{99325f1d-5f31-4eda-8b5a-8f20ff0afe09} \\
\texttt{11aa1ad7-4a0f-42c9-8b30-9c4337cb552c} & \texttt{5036cbf3-853e-41cc-a201-344d98a49f85} & \texttt{b531d592-e688-4b76-bf9d-1bec02d0433e} \\
\texttt{1a800193-c277-46d3-8c57-528e7e4fad16} & \texttt{5e3e89df-3bd0-476a-893a-2f0385878540} & \texttt{df0d24de-719f-479e-90f8-36f8544c61c1} \\
\texttt{1b4ae546-d557-44bd-81d0-263d4ee8f3a0} & \texttt{6930cda5-de6a-4706-96db-70a26bed8ed8} & \texttt{e041aa5e-3e1b-4644-885c-2d1aede10d93} \\
\texttt{3c896401-e3e3-4e5b-8c83-815b9823b0c6} & \texttt{6b697097-8b1f-45a0-ad9a-d754e139b15e} & \texttt{e6ca5d96-d289-42ba-a1e5-728b6a8417ed} \\
\texttt{3e1efc74-ad89-42ae-b87e-7ce7de5dcfa9} & \texttt{8d5e9ac9-8c08-40f3-b119-0c32717503cf} & \texttt{e6cc5cfa-50cb-442b-8c99-3d8931a97c0d} \\
\texttt{3f5497c6-ea3e-404c-93bb-64f19e6853fe} & \texttt{8f074e46-5f12-4ce7-824e-840fe4bb0884} & \texttt{edb8b137-3b93-4d76-8573-55444b42cf43} \\
\texttt{45bc3822-9054-4919-8bd9-7e87e24e67aa} & \texttt{98f59003-ba5e-46d9-91c0-e9a60118bc09} & \texttt{f6fbe4bb-162a-4c48-9ec5-05cb77231041} \\
\bottomrule
\end{tabular}
\caption{Identifiers of the 89 sessions that our tasks are built from. SWE-Together's tasks are named by their \texttt{task\_id} in its release, and SWE-chat's sessions by their \texttt{session\_id} in revision \texttt{202b071} of its release.}
\label{tab:task-ids}
\end{table*}

\paragraph{SWE-chat Sessions.}
To increase the number of tasks, we also select sessions from a newer release of SWE-chat that SWE-Together does not include.
We first apply the same filters as SWE-Together (e.g., a repository with at least 10 stars, at least 30\% of the code written by the agent), including its LLM-based check that a new agent can redo the work from the starting environment alone.
We then keep only sessions recorded with Claude Code or Codex that contain at least three messages from the user.
The starting commit of each session must also be recoverable from a public GitHub repository.
We exclude projects that would need a large amount of disk space to build (e.g., an Android SDK or Dart).
We also exclude sessions where the recorded agent relied on anything that is not available within the starting environment and requires external resources that may not be available during simulation (e.g., a GitHub issue or a tool that runs outside the repository).
Finally, as SWE-chat can include multiple sessions that come from the same repository, we increase diversity by selecting at most 8 sessions from each repository.
We create our own starting environments for each session that contains the repository at its starting commit and the runtime it needs.

\paragraph{Tasks Starting from Later Messages.}
In four of the 89 tasks, the session needs content outside of the starting environment (e.g., GitHub pages, external API) only in its first turns.
We start these tasks from a later user message that does not depend on that content.
The agent and the user simulator receive the earlier turns as conversation history.

\subsection{User Simulators}
\label{app:simulators}

All user simulators run on the same LLM (GPT-5.6 Luna) and write one reply after each agent turn.

\paragraph{SWE-Together.}
We use the simulator prompt that their released code provides for each task.
Some of our tasks exclude turns of the recorded session (e.g., the turns before the user message that a task starts from), but the provided prompt still describes them.
We remove these passages from the simulator prompt with an LLM (GPT-5.6 Terra).
For the SWE-chat tasks, which have no provided prompt, we use an LLM (Claude Opus 5.5) to write a prompt for each task in the form of the released prompts.

\paragraph{SWE-Interact and Goal-Prompted Simulator.}
In SWE-Interact, the simulated user runs in fixed stages (i.e., a set order in which it reveals requirements derived from benchmark tasks) and it is able to inspect the repository or environment through tool calls.
We take their prompt, but with the intents and criteria of our intent model in place of the benchmark requirements (Fig.~\ref{fig:prompt-swe-interact}).
Additionally, for a fair comparison with the other simulators, we prevent the simulator from using tool calls to inspect its repository and environment.
The goal-prompted simulator receives the same intents and criteria from our intent model with a plain instruction to reply as the user (Fig.~\ref{fig:prompt-simple-user}).

\subsection{Method}
\label{app:method}

This section gives the details of each test in Section~\ref{sec:method}.

\paragraph{Replication.}
\label{app:replication}
The LLM judge used to assess similarity of real user replies and simulated ones uses one prompt for single-turn replies and another one for the multi-turn setting (Figs.~\ref{fig:prompt-judge-reply} and~\ref{fig:prompt-judge-stretch}).
When we run the judge, we run it twice, average the two ratings, and rescale each rating to 0--100 as $100(r - 1) / 4$.
The recorded sessions of the 89 tasks contain 569 user replies that we used for evaluation.
At each reply, \framework{} starts from the intent state built from the conversation before the last agent turn, then assesses that turn, updates the intent state, and writes the reply.
For the multi-turn setting, we average the sets of three consecutive replies within each task, then average over the tasks that have at least one such set, which results in 71 tasks.

\paragraph{Robustness.}
\label{app:robustness}
We rewrite a session in two ways, merging consecutive user turns and paraphrasing messages.
For merging, we ask an LLM (GPT-5.6 Luna) whether each pair of consecutive user turns could have been sent as one message (Fig.~\ref{fig:prompt-merge-decide}).
For such identified turns, we use an LLM to merge the turns with a short connecting phrase (e.g., ``Also,''), which gives 153 merged turns over 72 tasks (Fig.~\ref{fig:prompt-merge-write}).
For paraphrasing, we use the LLM to paraphrase each user and agent message (Fig.~\ref{fig:prompt-reword}).
Then, for evaluation, we compare the simulated replies against the paraphrased versions of the real users' replies.

\paragraph{Task Goals.}
We use an LLM (GPT-5.6 Terra) to write the goals of each task from the recorded session alone, without the intent model (Fig.~\ref{fig:prompt-goal-writer}).
Each goal cites the user's messages that support it, and work that the recorded agent did on its own is left out.

\paragraph{Policies.}
\label{app:adaptation}
We use an LLM (GPT-5.6 Terra) to list, from the recorded session and its goals, every agent action that the user pushed back on (Fig.~\ref{fig:prompt-pushback-reader}).
We then use the same LLM to turn each action in this list into an observed instruction that repeats it in similar situations and a proactive instruction that avoids it (Fig.~\ref{fig:prompt-policy-writer}).
For example, consider an agent that was corrected for committing all staged changes before asking the user which ones to commit.
The observed instruction will tell the agent to always commit all staged changes, while its proactive counterpart would tell the agent to inspect staged changes first and confirm with the user which ones to commit.

\paragraph{Task Success.}
To score task success, an LLM judge (GPT-6.1 Sol) reads the goals, the changes that the replay made to the repository, and the agent's messages and tool calls (Fig.~\ref{fig:prompt-grader}).
The judge does not read the messages of the real or simulated user, so that it judges each replay on the agent's work alone.
We run the judge twice on each replay.
Success $s$ is the percentage of goals met, averaged over the two runs of the judge, and burden-adjusted success follows Section~\ref{sec:scoring}.
For the win rate, we compare each proactive-policy replay of a task with each of its observed-policy replays, counting a win as 1, a tie as 0.5, and a loss as 0, and average the four comparisons into the task's score~\citep{vargha2000critique}.
The win rate is the mean of these scores over tasks.
We test whether the proactive policy wins more tasks than it loses with a two-sided sign test, which leaves out tasks with a score of exactly 0.5, and apply a Holm correction across the five user simulators~\citep{holm1979simple}.

\subsection{Additional Results}
\label{app:additional-results}

This section gives further results for each dimension.

\paragraph{Replication and Robustness: Content and Style.}
\label{app:additional}
Table~\ref{tab:content-style} gives the content and style scores behind Table~\ref{tab:results}.
The style profile raises the style score of \framework{} in every setting, while its content score stays within 1.0 of the version without the profile.

\begin{table*}[t]
\centering
\small
\begin{tabular}{lcccc}
\toprule
& \multicolumn{2}{c}{\textbf{Replication}} & \multicolumn{2}{c}{\textbf{Robustness}} \\
\cmidrule(lr){2-3} \cmidrule(lr){4-5}
\textbf{User simulator} & \textbf{Single} & \textbf{Multi} & \textbf{Single} & \textbf{Multi} \\
\midrule
Goal-Prompted & 36.6 / 16.1 & 42.2 / 7.4 & 36.3 / 15.6 & 42.3 / 6.5 \\
SWE-Interact & 36.3 / 33.6 & 40.0 / 22.8 & 36.7 / 34.9 & 38.3 / 22.3 \\
SWE-Together & 49.0 / 53.7 & 39.1 / 41.7 & 47.2 / 49.0 & 40.6 / 41.9 \\
\midrule
\framework{} & 49.5 / 57.0 & 42.7 / 47.2 & 47.2 / 51.8 & 41.2 / 40.2 \\
\quad $-$ Style profile & 49.5 / 54.2 & 41.8 / 42.3 & 46.4 / 48.2 & 41.7 / 38.4 \\
\bottomrule
\end{tabular}
\caption{Content and style scores (content / style, scaled to 0 to 100) of each user simulator on the 89 tasks, judged by GPT-6.1 Sol. The mean of each pair is the similarity in Table~\ref{tab:results}.}
\label{tab:content-style}
\end{table*}

\paragraph{Session Length.}
We group the tasks by the number of user messages in the recorded session and compare each user simulator's scores across the groups (Table~\ref{tab:session-length}).
The scores of the goal-prompted simulator decrease as sessions get longer, SWE-Interact scores highest on the shortest sessions, and the scores of SWE-Together stay nearly the same.
One explanation is that SWE-Together's prompt lists all of the user's later messages, so it has the content of each reply however long the session is.
\framework{} scores about as high as SWE-Together in every group.

\begin{table}[t]
\centering
\small
\setlength{\tabcolsep}{3pt}
\begin{tabular}{lccc}
\toprule
\textbf{User messages} & \textbf{Up to 5} & \textbf{6 to 8} & \textbf{9 or more} \\
\midrule
\multicolumn{4}{@{}l}{\textit{Single-turn (39, 24, and 26 tasks)}} \\
Goal-Prompted & 32.9 & 26.4 & 24.0 \\
SWE-Interact & 37.1 & 34.4 & 36.0 \\
SWE-Together & 51.1 & 49.1 & 49.3 \\
\cmidrule{1-4}
\framework{} & 55.4 & 50.9 & 50.2 \\
\quad $-$ Style profile & 52.5 & 49.6 & 49.3 \\
\midrule
\multicolumn{4}{@{}l}{\textit{Multi-turn (22, 24, and 25 tasks)}} \\
Goal-Prompted & 28.8 & 24.6 & 22.0 \\
SWE-Interact & 33.1 & 30.2 & 30.7 \\
SWE-Together & 40.9 & 42.1 & 39.7 \\
\cmidrule{1-4}
\framework{} & 46.2 & 44.3 & 40.8 \\
\quad $-$ Style profile & 42.9 & 41.4 & 41.1 \\
\bottomrule
\end{tabular}
\caption{Similarity scores (scaled to 0 to 100) of each user simulator in replication and robustness by the number of user messages in the recorded session. Each task's score is the mean of its recorded and rewritten sessions. The multi-turn setting needs longer sessions, so this part has fewer tasks.}
\label{tab:session-length}
\end{table}

\begin{figure*}[t]
\centering
\includegraphics[width=\textwidth]{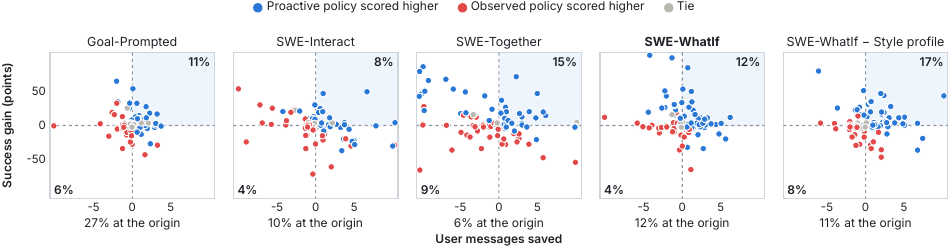}
\caption{Difference between the proactive-policy and observed-policy replays of each task, per user simulator. The x-axis gives how many fewer messages the simulated user sent with the proactive policy, and the y-axis how many more percentage points of goals the proactive-policy replay met. Color shows which replay had the higher burden-adjusted success, and values beyond the axes are drawn at the edge. The percentages give the share of tasks in the upper right (proactive policy ahead on both), in the lower left (behind on both), and at the origin (no difference).}
\label{fig:replay-differences}
\end{figure*}

\paragraph{Differences between Replays.}
Figure~\ref{fig:replay-differences} compares the two replays of each task by mapping their difference in terms of success against their difference in terms of the number of messages that the simulated user sent.
With a simulator that adapts to the agent, the proactive-policy replay should reach higher success, need fewer messages, or both.
With SWE-Together, the proactive-policy replay reached higher success but needed more messages in 25\% of tasks, three or more in most of them (top-left quadrant in its Figure).
As seen in the figure, for the goal-prompted simulator, the replays with both policies were generally similar, where they are all grouped in the middle.
In contrast, with \framework{} the proactive-policy replay needed fewer messages in 43\% of tasks with the style profile and 52\% without.

\begin{figure}[t]
\centering
\includegraphics[width=\columnwidth]{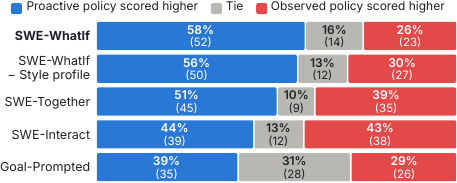}
\caption{Share of the 89 tasks in which the proactive-policy replay achieved higher burden-adjusted success than the observed-policy replay (win), the same (tie), or lower (loss), per user simulator, with the number of tasks in parentheses.}
\label{fig:win-loss}
\end{figure}

Figure~\ref{fig:win-loss} shows the share of tasks in which each user simulator's proactive-policy replay achieved higher, equal, or lower burden-adjusted success than its observed-policy replay.

\paragraph{Error Analysis.}
\label{app:error-annotation}
Table~\ref{tab:error-kinds} defines the four kinds of error used in the analysis in Section~\ref{sec:results}.
One author manually labeled simulated turns from 100 replays by comparing them to the recorded sessions, randomly sampled across tasks and user simulators.
Then, the inter-rater agreement on these labels with Claude Sonnet 5.5 was calculated, which provided a Cohen's $\kappa$ of 0.652.
As this denotes substantial agreement, the same model was used to annotate the turns of the first replay with each policy in every task of the adaptation test.
We counted the number of replays that were annotated to contain an error to produce the frequency results (Table~\ref{tab:errors}).

\begin{table}[t]
\centering
\small
\begin{tabular}{p{0.17\linewidth}p{0.71\linewidth}}
\toprule
\textbf{Error} & \textbf{What the simulated user does} \\
\midrule
Premature & Asks for something before the point where the real user asked for it, or bundles requests that the real user made one at a time. \\
Extra & Asks for work that the real user never asked for (e.g., diffs, tests, commits, or approvals). \\
Passive & Lets the agent keep working where the real user stepped in with feedback or a correction, or keeps the session going after the real user stopped. \\
Omitted & Leaves out a request or detail of the real user that still applies to the agent's work. A correction for a mistake that the agent did not make is not counted. \\
\bottomrule
\end{tabular}
\caption{Types of error that were made by the user simulators during the simulated replays for the adaptation test.}
\label{tab:error-kinds}
\end{table}

\begin{figure*}[t]
\centering
\includegraphics[width=\textwidth]{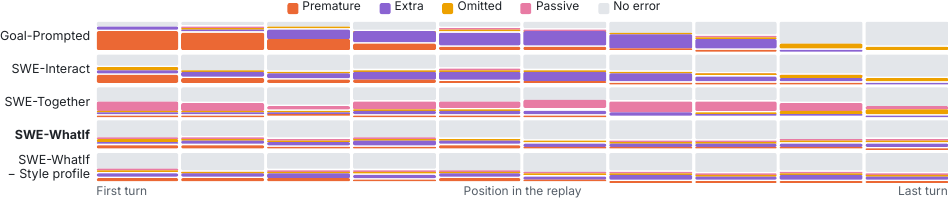}
\caption{Share of each user simulator's turns with each kind of error, by position in the replay. Each replay's turns are divided into ten equal buckets from its first turn to its last.}
\label{fig:error-timing}
\end{figure*}

\paragraph{Errors by Position.}
Figure~\ref{fig:error-timing} shows where in a replay each kind of error occurred, where we divide each replay into ten equal buckets based on how close they are to the start or end of their respective sessions. 
As seen, the goal-prompted simulator tended to provide premature requests in the first half of its replays, and SWE-Interact provides extra requests in the middle of the session.
SWE-Together tended to provide passive replies throughout the whole session, demonstrating how the simulated trajectory could diverge from the recorded one throughout a session.
\framework{}, in general, had fewer errors, which were also spread across the replay.

\paragraph{Turing Test.}
\label{app:turing}
Following SimulatorArena~\citep{dou2025simulatorarena}, we used an LLM judge (GPT-6 Luna) to test whether each user simulator could be distinguished from the real user.
The judge was given two versions of the same interaction, one with the real user and one with the simulated user, and decided which one involved the real user.
We perform this test at the turn-level and the session-level.
For the turn-level, we use the simulated replies and corresponding recorded replies for the replication test.
For the session-level, we use the recorded sessions and the replays performed with the observed-policy for the adaptation evaluation.
Due to the positional bias of LLMs, we ran each comparison twice with the versions in each position, and used the judge's confidence to break the tie when it chose the same position both times.
We report the Turing distance, $|p - 50|$, where $p$ is the percentage of comparisons in which the judge identified the real user.
As seen from the results Table~\ref{tab:turing}, both versions of \framework{} had a lower Turing distance than the other user simulators at both levels.

\begin{table}[t]
\centering
\small
\begin{tabular}{lcc}
\toprule
\textbf{User Simulator} & \textbf{Turn-Level} & \textbf{Session-Level} \\
\midrule
Goal-Prompted & 42.2 & 44.4 \\
SWE-Interact & 26.0 & 28.7 \\
SWE-Together & 26.3 & 47.8 \\
\midrule
\textbf{\framework{}} & \underline{19.7} & \textbf{24.7} \\
\quad $-$ Style profile & \textbf{19.3} & \textbf{24.7} \\
\bottomrule
\end{tabular}
\caption{Turing distance of each user simulator on single turns and whole sessions, judged by GPT-6 Luna. Lower is better, and 0 means that the judge cannot distinguish the simulated user from the real user. Best results are \textbf{bold}-faced and second best results are \underline{underlined}.}
\label{tab:turing}
\end{table}

\subsection{Evaluating LLMs as Coding Agents}
\label{app:agents}

\begin{figure}[t]
\centering
\includegraphics[width=\columnwidth]{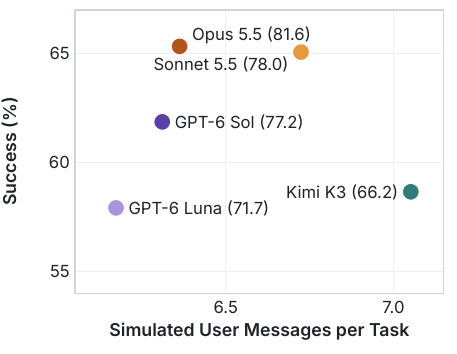}
\caption{Mean success of each model against the mean number of messages that \framework{} sent per task, with mean burden-adjusted success in brackets.}
\label{fig:success-messages}
\end{figure}

We replay 58 of the 65 tasks from SWE-Together's sessions once with each of the five models, where we left out seven tasks in which an agent could not complete a turn within its time limit of 40 minutes.

\paragraph{Setup.}
We use opencode~\citep{anomaly2025opencode} as the harness with Claude Opus 5.5, Claude Sonnet 5.5, GPT-6 Sol, GPT-6 Luna, and Kimi K3.
The agents use Amazon Bedrock with opencode's High reasoning setting.
The user simulator used was \framework{} with the style profile, and each model's mean burden-adjusted success over all replays gives its rank.

\paragraph{Published Rankings.}
We compare our ranking with three published rankings (Table~\ref{tab:published-rankings}).
SWE-Together's leaderboard~\citep{wu2026swe} runs each model in opencode with its simulated user, and orders the models by pass@1 (i.e., the percentage of tasks that the agent solves in its first attempt).
For Agent Arena~\citep{arena2026agent}, we use its leaderboard for the Code category of October 8, 2026, which ranks models from real users' sessions with coding agents (e.g., whether the task was completed) by their \emph{net improvement}.
This leaderboard lists Claude Opus 5.5 at the High reasoning setting and the other four models at Max.
For a ranking without users, we use the DeepSWE~\citep{huang2026deepswe} scores that Artificial Analysis measured for its Coding Agent Index~\citep{aa2026codingagent}, with each model at its highest reasoning setting in its developer's own harness (Claude Code, Codex, or Kimi Code CLI).
We measure agreement between two rankings with Kendall's $\tau_b$, which accounts for ties.
We read the leaderboards on October 11, 2026.

\begin{table}[t]
\centering
\footnotesize
\setlength{\tabcolsep}{1.2pt}
\begin{tabular}{lcccccc}
\toprule
& \multicolumn{3}{c}{\textbf{\framework{}}} & & & \\
\cmidrule(lr){2-4}
\textbf{Model} & \textbf{Succ.} & \textbf{Msgs.} & \textbf{Score} & \textbf{DeepSWE} & \textbf{SWE-T.} & \textbf{Arena} \\
\midrule
Opus 5.5 & 65.3 & 6.4 & 81.6 & 68.4 & 61.5 & 17.0 \\
Sonnet 5.5 & 65.1 & 6.7 & 78.0 & 72.0 & 51.4 & 12.2 \\
GPT-6 Sol & 61.9 & 6.3 & 77.2 & 69.0 & 50.5 & 10.7 \\
GPT-6 Luna & 57.9 & 6.2 & 71.7 & 63.7 & 47.7 & 2.8 \\
Kimi K3 & 58.7 & 7.1 & 66.2 & 68.4 & 53.2 & 3.9 \\
\midrule
$\tau_b$ with Arena & & & .80 & .53 & .60 & 1 \\
\bottomrule
\end{tabular}
\caption{Scores of the five models. For \framework{}, success (Succ.) is the mean percentage of goals met, Msgs.\ is the mean number of messages that the simulated user sent, and Score is the mean burden-adjusted success, all over the 58 tasks. DeepSWE gives Artificial Analysis's score in percent, SWE-Together's leaderboard (SWE-T.) gives pass@1 in percent, and Agent Arena (Arena) gives the net improvement in percent for the Code category. The last row gives the agreement of each ranking with Agent Arena's.}
\label{tab:published-rankings}
\end{table}

\paragraph{User Groups.}
We group the tasks by two properties of their intent models (Table~\ref{tab:agent-groups}).
For \emph{reactive intents}, we separate the 23 tasks with at least one intent that the user raises only if a specific situation arises during the replay from the 35 tasks without such an intent.
For the \emph{opening request}, we separate the 26 detailed tasks that state more than two criteria of their intents from the 32 brief tasks that state at most two.

\begin{table*}[t]
\centering
\footnotesize
\setlength{\tabcolsep}{4.0pt}
\begin{tabular}{lcccccc}
\toprule
& \multicolumn{3}{c}{\textbf{Without Reactive Intents ($n=35$)}} & \multicolumn{3}{c}{\textbf{With Reactive Intents ($n=23$)}} \\
\cmidrule(lr){2-4} \cmidrule(lr){5-7}
\textbf{Model} & \textbf{Succ.} & \textbf{Msgs.} & \textbf{Score} & \textbf{Succ.} & \textbf{Msgs.} & \textbf{Score} \\
\midrule
Opus 5.5 & 71.9 & 5.6 & 79.4 & 55.4 & 7.5 & 85.0 \\
Sonnet 5.5 & 71.6 & 6.7 & 70.6 & 55.1 & 6.8 & 89.2 \\
GPT-6 Sol & 71.9 & 5.5 & 86.3 & 46.6 & 7.5 & 63.2 \\
GPT-6 Luna & 67.5 & 5.7 & 76.9 & 43.4 & 6.9 & 63.7 \\
Kimi K3 & 63.6 & 6.0 & 68.2 & 51.1 & 8.7 & 63.0 \\
\midrule
& \multicolumn{3}{c}{\textbf{Detailed Opening Request ($n=26$)}} & \multicolumn{3}{c}{\textbf{Brief Opening Request ($n=32$)}} \\
\cmidrule(lr){2-4} \cmidrule(lr){5-7}
\textbf{Model} & \textbf{Succ.} & \textbf{Msgs.} & \textbf{Score} & \textbf{Succ.} & \textbf{Msgs.} & \textbf{Score} \\
\midrule
Opus 5.5 & 65.1 & 7.7 & 70.5 & 65.5 & 5.2 & 90.7 \\
Sonnet 5.5 & 65.6 & 8.4 & 65.8 & 64.6 & 5.4 & 87.9 \\
GPT-6 Sol & 63.5 & 7.4 & 78.1 & 60.5 & 5.4 & 76.4 \\
GPT-6 Luna & 61.8 & 7.3 & 74.5 & 54.7 & 5.3 & 69.4 \\
Kimi K3 & 58.1 & 7.7 & 64.3 & 59.1 & 6.5 & 67.7 \\
\bottomrule
\end{tabular}
\caption{Performance within each group of tasks. Success (Succ.) is the mean percentage of goals met, Msgs.\ is the mean number of messages that the simulated user sent, and Score is mean burden-adjusted success.}
\label{tab:agent-groups}
\end{table*}

\section{User Study Details}
\label{app:user-study}
\label{app:study}

This appendix gives the details of the user study in Section~\ref{sec:user-study}.

\begin{figure*}[t]
\centering
\begin{minipage}[t]{0.49\textwidth}\vspace{0pt}\centering\includegraphics[width=\linewidth]{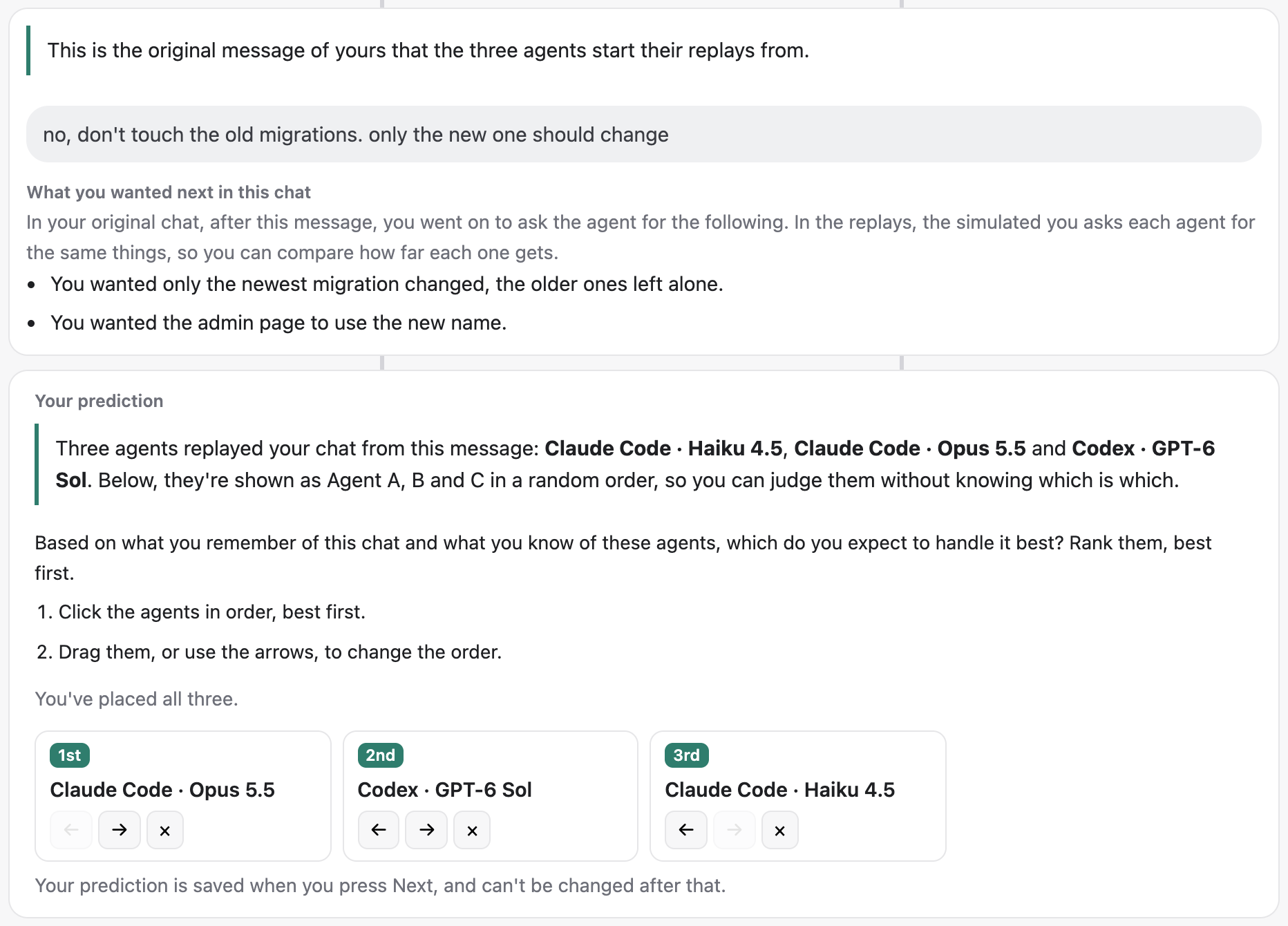}\\(a) Prediction\end{minipage}\hfill
\begin{minipage}[t]{0.49\textwidth}\vspace{0pt}\centering\includegraphics[width=\linewidth]{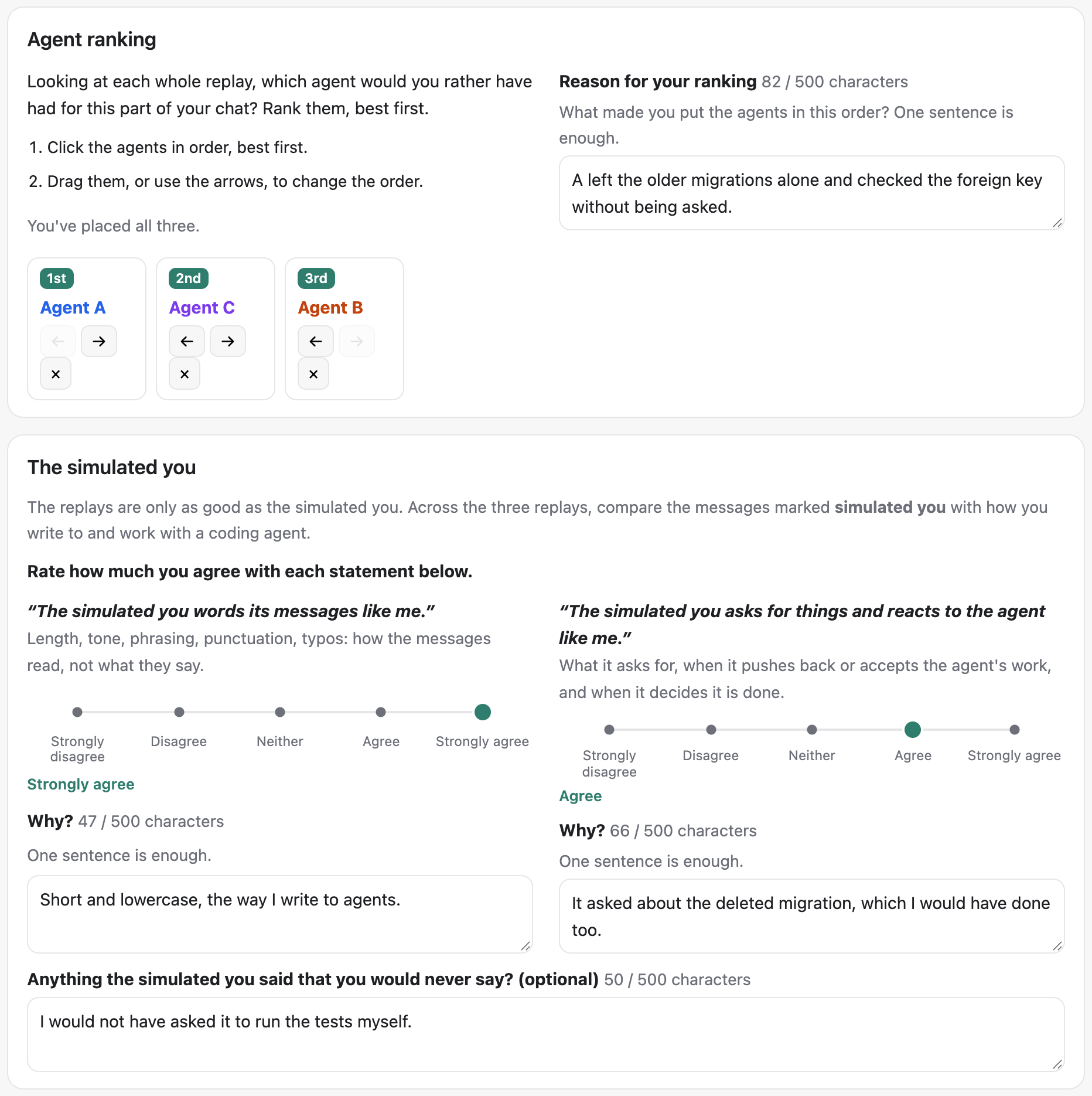}\\(c) Final ranking and ratings\end{minipage}
\par\medskip
\includegraphics[width=\textwidth]{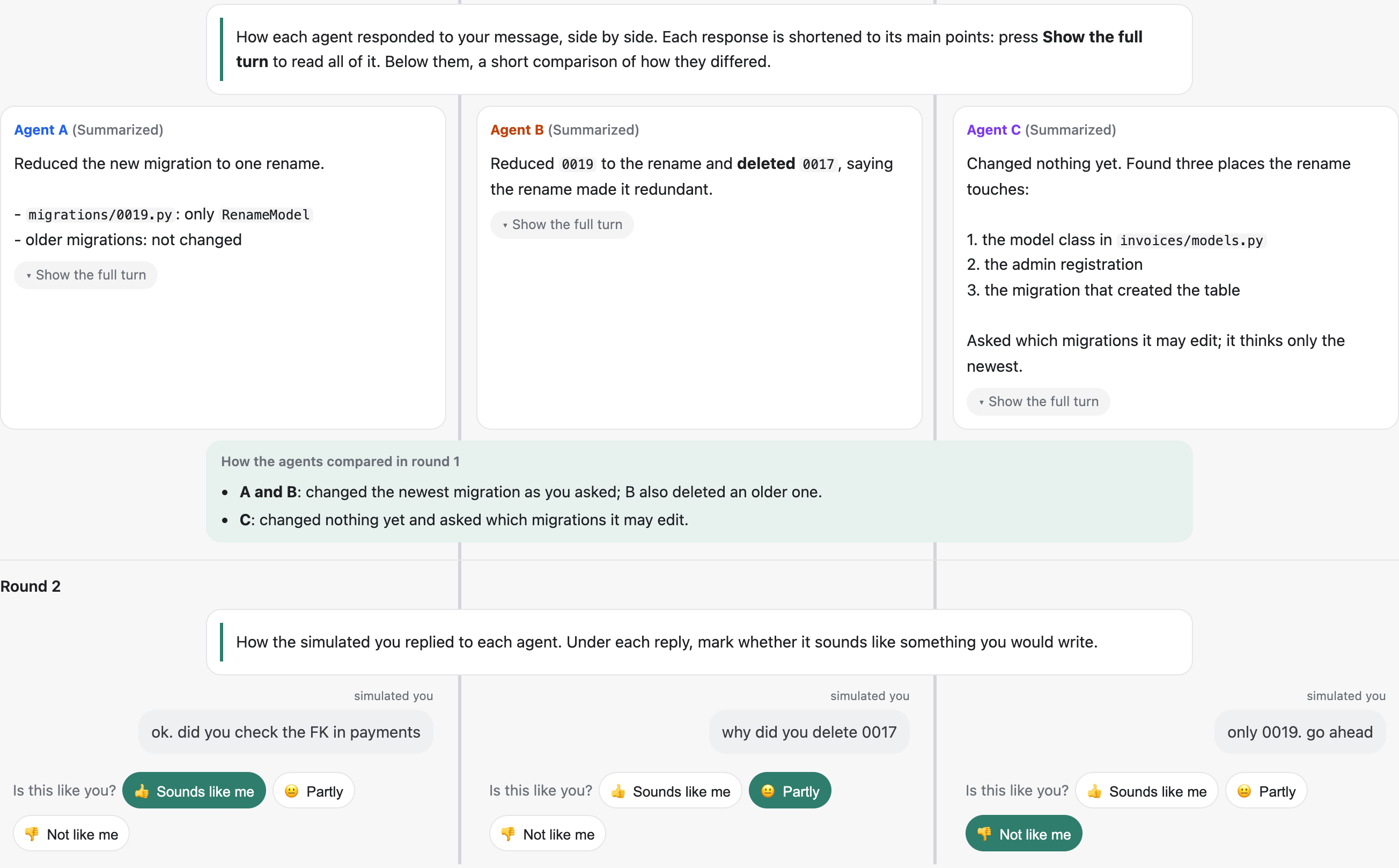}\\(b) One turn of the replays
\caption{Interface used in the user study, where participants (1) selected their recorded sessions, (2) chose one session, reviewed the sensitive details that we replaced in it, gave consent and provided it, (3) answered background questions and waited while the replays ran, (4) read a summary of the session up to the message from which the replays start and predicted how the three named agents would rank, (5) read the three replays side by side one turn at a time with the agents hidden and marked each reply of the simulated user, (6) ranked the agents and rated the simulated user, and (7) saw which agent was which and answered two questions about what they learned.}
\label{fig:study-interface}
\end{figure*}

\paragraph{Recruitment and Pay.}
We recruited participants through Prolific who were 18 or older, had recorded Claude Code or Codex sessions on their own computer, had an approval rate of at least 95\%, at least 100 previous submissions, English as their primary language, and residence in the United Kingdom or the United States.
We screened out participants whose recorded sessions had no user message from which a replay could start.
Participants were paid \pounds 11 for about 70 minutes, including a wait of 20 to 40 minutes while the replays ran, \pounds 0.50 if screened out, and \pounds 3 if they withdrew during the study or their replays could not be run.
Participants took 51 minutes on average, so the pay of \pounds 11 was above the \pounds 9 per hour recommended by Prolific.

\paragraph{Interface.}
Figure~\ref{fig:study-interface} shows the study interface, filled with fictional sessions in place of a participant's.
Above the replays, participants read their first message, a summary of the session before the user message from which the three replays start, that message itself, and a summary of their goals for the session.
Each agent turn was shown as a short summary, with the full turn available on request.
We generated all summaries with GPT-6 Luna.
The three replays were labelled anonymously and in a random order.

\begin{figure}[t]
\centering
\definecolor{sdis}{HTML}{E34948}
\definecolor{dis}{HTML}{F2A4A3}
\definecolor{neu}{HTML}{DCDBD7}
\definecolor{agr}{HTML}{9CC1EE}
\definecolor{sagr}{HTML}{2A78D6}
\begin{tikzpicture}[x=1cm,y=1cm,font=\footnotesize]
\node[anchor=south west,inner sep=0] at (0,0.00) {Words its messages like me};
\node[anchor=east,inner sep=0] at (1.8,-0.24) {Claude Code (7)};
\fill[dis] (1.920,-0.40) rectangle (2.466,-0.08);
\node[text=black,inner sep=0] at (2.193,-0.24) {1};
\fill[neu] (2.506,-0.40) rectangle (3.051,-0.08);
\node[text=black,inner sep=0] at (2.779,-0.24) {1};
\fill[agr] (3.091,-0.40) rectangle (5.980,-0.08);
\node[text=black,inner sep=0] at (4.536,-0.24) {5};
\node[anchor=east,inner sep=0] at (1.8,-0.62) {Codex (9)};
\fill[dis] (1.920,-0.78) rectangle (2.336,-0.46);
\node[text=black,inner sep=0] at (2.128,-0.62) {1};
\fill[agr] (2.376,-0.78) rectangle (5.980,-0.46);
\node[text=black,inner sep=0] at (4.178,-0.62) {8};
\node[anchor=south west,inner sep=0] at (0,-1.26) {Asks for things and reacts like me};
\node[anchor=east,inner sep=0] at (1.8,-1.50) {Claude Code (7)};
\fill[dis] (1.920,-1.66) rectangle (2.466,-1.34);
\node[text=black,inner sep=0] at (2.193,-1.50) {1};
\fill[agr] (2.506,-1.66) rectangle (5.980,-1.34);
\node[text=black,inner sep=0] at (4.243,-1.50) {6};
\node[anchor=east,inner sep=0] at (1.8,-1.88) {Codex (9)};
\fill[dis] (1.920,-2.04) rectangle (2.336,-1.72);
\node[text=black,inner sep=0] at (2.128,-1.88) {1};
\fill[neu] (2.376,-2.04) rectangle (2.791,-1.72);
\node[text=black,inner sep=0] at (2.583,-1.88) {1};
\fill[agr] (2.831,-2.04) rectangle (5.524,-1.72);
\node[text=black,inner sep=0] at (4.178,-1.88) {6};
\fill[sagr] (5.564,-2.04) rectangle (5.980,-1.72);
\node[text=white,inner sep=0] at (5.772,-1.88) {1};
\node[anchor=south west,inner sep=0] at (0,-2.52) {Taught me something about the agents};
\node[anchor=east,inner sep=0] at (1.8,-2.76) {Claude Code (7)};
\fill[agr] (1.920,-2.92) rectangle (4.809,-2.60);
\node[text=black,inner sep=0] at (3.364,-2.76) {5};
\fill[sagr] (4.849,-2.92) rectangle (5.980,-2.60);
\node[text=white,inner sep=0] at (5.414,-2.76) {2};
\node[anchor=east,inner sep=0] at (1.8,-3.14) {Codex (9)};
\fill[neu] (1.920,-3.30) rectangle (2.336,-2.98);
\node[text=black,inner sep=0] at (2.128,-3.14) {1};
\fill[agr] (2.376,-3.30) rectangle (5.069,-2.98);
\node[text=black,inner sep=0] at (3.722,-3.14) {6};
\fill[sagr] (5.109,-3.30) rectangle (5.980,-2.98);
\node[text=white,inner sep=0] at (5.544,-3.14) {2};
\node[anchor=south west,inner sep=0] at (0,-3.78) {Will help me decide how to use agents};
\node[anchor=east,inner sep=0] at (1.8,-4.02) {Claude Code (7)};
\fill[neu] (1.920,-4.18) rectangle (2.466,-3.86);
\node[text=black,inner sep=0] at (2.193,-4.02) {1};
\fill[agr] (2.506,-4.18) rectangle (4.223,-3.86);
\node[text=black,inner sep=0] at (3.364,-4.02) {3};
\fill[sagr] (4.263,-4.18) rectangle (5.980,-3.86);
\node[text=white,inner sep=0] at (5.121,-4.02) {3};
\node[anchor=east,inner sep=0] at (1.8,-4.40) {Codex (9)};
\fill[dis] (1.920,-4.56) rectangle (2.336,-4.24);
\node[text=black,inner sep=0] at (2.128,-4.40) {1};
\fill[agr] (2.376,-4.56) rectangle (5.524,-4.24);
\node[text=black,inner sep=0] at (3.950,-4.40) {7};
\fill[sagr] (5.564,-4.56) rectangle (5.980,-4.24);
\node[text=white,inner sep=0] at (5.772,-4.40) {1};
\node[anchor=west,inner sep=0,font=\scriptsize] at (0,-5.09) {\tikz[baseline=-0.6ex]\fill[sdis] (0,-0.09) rectangle (0.18,0.09);\,Strongly disagree\enspace \tikz[baseline=-0.6ex]\fill[dis] (0,-0.09) rectangle (0.18,0.09);\,Disagree\enspace \tikz[baseline=-0.6ex]\fill[neu] (0,-0.09) rectangle (0.18,0.09);\,Neutral\enspace \tikz[baseline=-0.6ex]\fill[agr] (0,-0.09) rectangle (0.18,0.09);\,Agree\enspace \tikz[baseline=-0.6ex]\fill[sagr] (0,-0.09) rectangle (0.18,0.09);\,Strongly agree};
\end{tikzpicture}
\caption{Participants' ratings of their simulated selves (top two) and of what they learned from the replays (bottom two), by the harness of their session. Numbers inside the bars are counts of participants.}
\label{fig:study-ratings}
\end{figure}
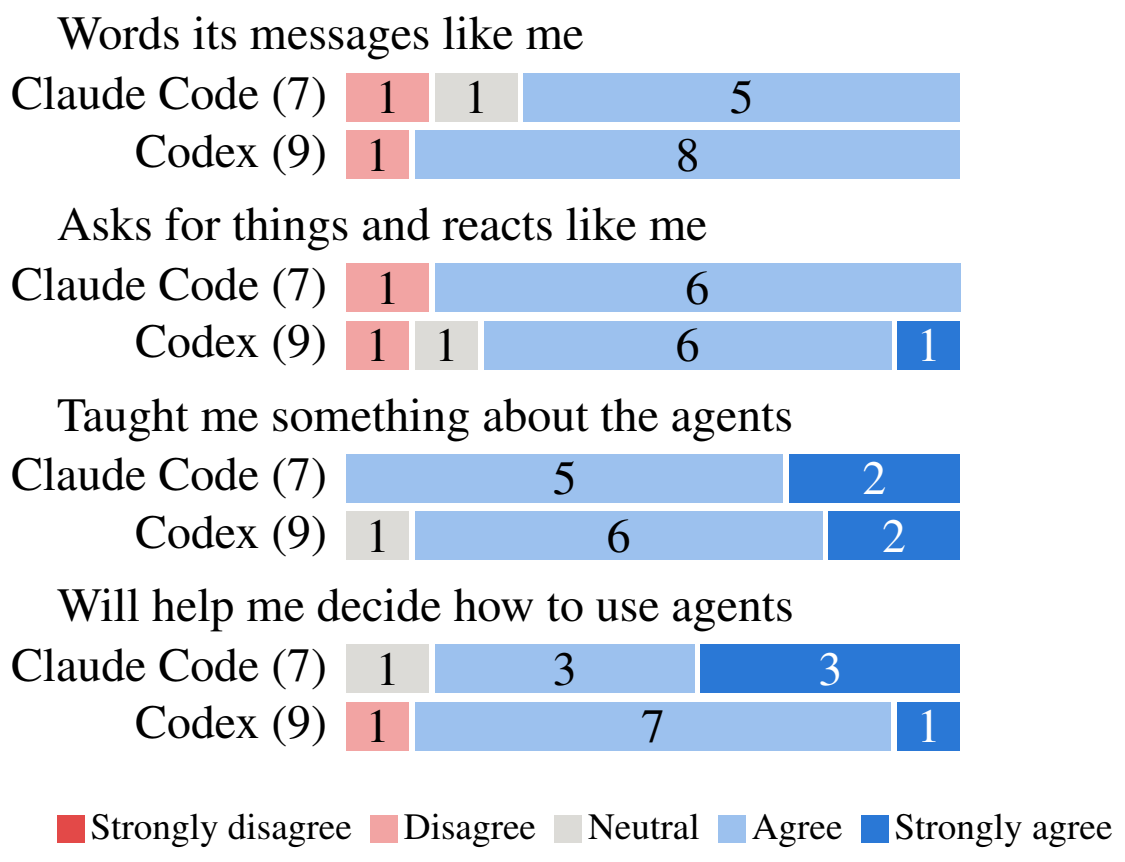

\begin{figure*}[t]
\centering
\includegraphics[width=\textwidth]{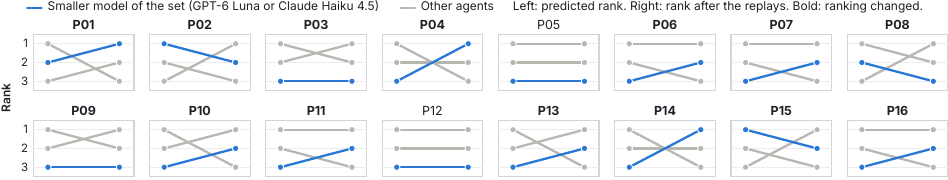}
\caption{Each participant's ranking of the three agents before reading the replays (left) and after (right), where 1 is best. The smaller model of each set (GPT-6 Luna or Claude Haiku 4.5) is drawn in blue. Participants who changed their ranking are in bold.}
\label{fig:study-rank-change}
\end{figure*}

\paragraph{Replay Start.}
For the start of each replay, we first identified user messages in each recorded session that were followed by at least three more exchanges (i.e., a user message and the agent's response).
Among these messages, we used GPT-5.6 Luna to judge whether the participant had pushed back on the agent at each of these messages (Fig.~\ref{fig:prompt-study-pushback}).
If a pushback was found, the replays started from the earliest of these messages with pushback, and, if none was found, simply from the earliest of these messages.
For 6 of the 16 analysed participants, the replays started from a user message where the participant had pushed back on the agent.
A recorded session does not include its repository, so we reconstructed the starting environment by extracting the content of all files that the recorded agent read or wrote from the recording itself.
The starting environment had no installed dependencies, no network access, and no running services, and we told every agent so.

\paragraph{User Simulator for Replay.}
We built each intent model from the participant's recorded session and its reconstructed starting environment, with the same drafting, reviewing, and revising agents as in our experiments (Appendix~\ref{app:construction}).
The simulated user was the \framework{} user simulator, with its style profile.
A replay stopped after three replies from the simulated user, or earlier once every intent was satisfied or dropped.

\paragraph{Agents.}
Each participant saw three agents, chosen by the harness that was used for the recorded session.
If participants were Claude Code users, they saw replays with Claude Opus 5.5 and Claude Haiku 4.5 in Claude Code and GPT-6 Sol in Codex.
If they were Codex users, they saw replays with GPT-6 Sol and GPT-6 Luna in Codex and Claude Sonnet 5.5 in Claude Code.
Of the 16 analysed participants, 7 provided a Claude Code session and 9 a Codex session.

\paragraph{Questionnaire.}
Participants answered the following items in order.
The agreement statements used a 5-point scale from ``Strongly disagree'' to ``Strongly agree''.
\begin{itemize}
    \item Prediction, with the agents named: ``Based on what you remember of this chat and what you know of these agents, which do you expect to handle it best? Rank them, best first.''
    \item Each reply of the simulated user, with the agents hidden: ``Sounds like me'', ``Partly'' or ``Not like me''.
    \item Final ranking: ``Looking at each whole replay, which agent would you rather have had for this part of your chat? Rank them, best first.'' with a written reason.
    \item ``The simulated you words its messages like me.'' (agreement), with ``Why?''
    \item ``The simulated you asks for things and reacts to the agent like me.'' (agreement), with ``Why?''
    \item After the agents were revealed: ``Seeing these three replays taught me something about these agents that I didn't know.'' (agreement), with ``What did you learn?''
    \item ``What I saw here will help me decide how to use coding agents from now on.'' (agreement), with ``What, if anything, will you do differently in the future?''
\end{itemize}

\begin{table}[t]
\centering
\footnotesize
\setlength{\tabcolsep}{4pt}
\begin{tabular}{lcc}
\toprule
\textbf{Agent} & \textbf{Predicted} & \textbf{Final} \\
\midrule
\multicolumn{3}{@{}l}{\textit{Claude Code sessions (7 participants)}} \\
Claude Opus 5.5 & 6 & 3 \\
Claude Haiku 4.5 & 0 & 2 \\
GPT-6 Sol (Codex) & 1 & 2 \\
\midrule
\multicolumn{3}{@{}l}{\textit{Codex sessions (9 participants)}} \\
GPT-6 Sol & 4 & 4 \\
GPT-6 Luna & 2 & 1 \\
Claude Sonnet 5.5 (Claude Code) & 3 & 4 \\
\bottomrule
\end{tabular}
\caption{Number of participants who ranked each agent first in their prediction and in their final ranking, by the harness of their recorded session.}
\label{tab:study-first}
\end{table}

\begin{table*}[t]
\centering
\footnotesize
\begin{tabular}{@{}p{3.4cm}p{5.4cm}cp{5.2cm}@{}}
\toprule
\textbf{Theme} & \textbf{Description} & $\mathbf{n}$ & \textbf{Example} \\
\midrule
\multicolumn{4}{@{}l}{\textit{What participants learned}} \\
Smaller models were enough & A smaller and cheaper model met the same goals as the larger one, or did better. & 12 & ``can still do decent investigative work'' (P11, on Claude Haiku 4.5) \\
Agents worked differently with the user & Agents differed in how they asked, listened, and explained, and this affected which one the participant preferred. & 8 & ``felt the most thorough, the most like a team member with competence'' (P13) \\
Beliefs about harnesses were tested & The replays confirmed or revised which harness the participant expected to be better. & 4 & ``Claude Code still outdoes ChatGPT.'' (P12) \\
How agents are steered matters & What the agent is told, and how early, shapes the result. & 2 & ``All three needed more context for a reliable calculation.'' (P05) \\
\midrule
\multicolumn{4}{@{}l}{\textit{What participants will do differently}} \\
Choose a model for each task & Start with a smaller model, match the model to the task, or compare models before choosing. & 11 & ``I can move between models when I need to level up or down the response'' (P15) \\
Steer agents differently & Give context earlier or change instructions, based on what the replays showed. & 4 & ``give the agent the source figures and assumptions early'' (P05) \\
Keep the current harness & Keep the harness they use, as the replays confirmed it. & 3 & ``I still find it by far the best'' (P16) \\
\midrule
\multicolumn{4}{@{}l}{\textit{Where the simulated user differed}} \\
More formal or wordier & The simulated user wrote more formally, lengthily, or completely than the participant. & 5 & ``a bit wordier than I would normally write'' (P11) \\
Kept insisting on the same thing & The simulated user kept asking for the same thing, when the participant felt that they would have not. & 3 & kept asking for a live test ``after it already explained it lacked network and shell access'' (P09) \\
\bottomrule
\end{tabular}
\caption{Themes in participants' written answers, with the number of participants ($n$) and one example each. A participant can fit more than one theme.}
\label{tab:study-themes}
\end{table*}

\paragraph{Participants.}
Participants answered six background questions after providing their session.
Of the 16 analysed participants, 4 were software engineers, 2 researchers, 1 a data scientist or ML engineer, 1 a student, 2 in other technical roles and 6 in non-technical roles.
Seven had programmed for less than three years, 6 for three to five years, and 3 for six years or more.
Seven had used coding agents for less than a year and 9 for a year or more.
Eight used coding agents every day or almost every day, 4 several times a week, 3 about once a week, and 1 less often.
Eleven used Claude Code regularly and 11 used Codex, and 6 used both.
Eight used them mostly for personal projects, 3 mostly for work, and 5 for both about equally.

\paragraph{Privacy.}
Before the session was sent to us, we replaced sensitive details (e.g., API keys, user names in file paths, email addresses and names in git author lines) with placeholders on the participant's computer, and showed the participant each replacement.
We repeated the replacement after their sessions were uploaded to our server and the interface only uploaded the part of each session up to three exchanges after the last message from which a replay could start.
Once the replays were ready, we deleted the participant's session and the intent model, style profile, and starting environment built from it.

\paragraph{Additional Results.}
Figure~\ref{fig:study-ratings} shows the distribution of participants' ratings to the four questions by which agent tool they use.
Table~\ref{tab:study-first} shows which agent participants ranked first before and after reading the replays.
Figure~\ref{fig:study-rank-change} shows each participant's ranking of the three agents before and after reading the replays.
Table~\ref{tab:study-themes} lists the themes in participants' written answers, with one example each.
Claude Code users marked 51\% of the simulated replies as ``Sounds like me'', 34\% as ``Partly'', and 15\% as ``Not like me'', and Codex users 48\%, 42\%, and 10\%.

\subsection{Participant Information and Consent}
\label{app:participant-instructions}

The following text reproduces the study overview, session-selection instructions, and consent agreement shown in the study interface.
Researcher names, institutional details, contact information, and the approval reference are replaced with bracketed descriptions for anonymous review.
The questionnaire and replay interface are given earlier in this appendix.

\subsubsection{Study Overview}
\paragraph{What happens next}
Plan for about 70 minutes in all: about 30 minutes of your own time, and a 20 to 40 minute wait while the replays run. During the wait you can do something else: leave the study tab open in the background and it notifies you when the replays are ready, or close it and come back with your results link.\par
Consent (about 2 minutes). Read what we collect, who reads it and how long we keep it, and agree.\par
Choose chats (about 3 minutes). See exactly what would be uploaded from each usable chat, and choose which to upload to our server.\par
Background questions, then a wait (about 2 minutes, then 20 to 40 minutes) while we create the simulated you and run the three replays.\par
Compare the replays (about 20 minutes). Predict which agent will do best, read the replays side by side and mark each message from the simulated you, then rank the agents and rate the simulated you.\par
See which agent was which (about 3 minutes), beside your prediction, and answer two short questions.\par
We pay you \pounds 11 through Prolific when you finish. The consent page lists every payment.\par
\paragraph{What we collect}
Your chats are uploaded to our server only to create the simulated you and run the three replays. Once the replays are ready, we delete your chats and the simulated you.\par
From the chat we use, we keep your first message, a short summary of what happened before the message where the agents start, that message, a short summary of what you wanted from the replayed part, and the three replays (what the agents and the simulated you wrote, and what the agents did in the copy of your project).\par
Your prediction, your marks on the simulated you's replies, your ratings, your ranking and your written responses.\par
Your answers to the background questions.\par
Your Prolific ID.\par
We use all of this only to analyse the study's results. It is never released as a dataset or shared outside the research team, apart from the AI model providers that run the replays. We publish only the results, such as figures, and short excerpts with nothing that identifies you.\par
The consent page says who reads this and how long we keep it.\par

\subsubsection{Session Selection}
Select each chat you're comfortable uploading to our server. We replay one of them. After you upload, our server checks your chats once more. If it can't use the first, it tries the next, so uploading more than one helps. Chats we don't use are deleted once the replays are ready.\par
\paragraph{What would be uploaded}
Only part of each chat: everything from its start up to a few exchanges after the message where the replay would start. An exchange is one of your messages and the agent's reply. In a short chat, that can be all of it.\par
Before anything is uploaded, our website deletes sensitive details from your chats, including keys, passwords, your user name and email addresses, and puts a placeholder in their place.\par
Each chat below shows its first message and the sensitive details that will be deleted from it. Press Read what would be uploaded to see the full text of the chat before you select a chat.\par

\subsubsection{Consent Agreement}
\paragraph{About this study}
[Study title], run by [researchers] at the [institution], and approved by the [institutional review board] ([approval reference]). This study replays part of one of your chats with three different coding agents, using a simulated version of you that writes to each agent in your place. It asks which agent you would rather have had, and how closely the simulated you resembles you. The parts below say what you would send, what happens to it, and your rights.\par
\paragraph{1. What is uploaded, and why}
From each chat you choose to upload: everything from its start up to three exchanges after the last message of yours the agents could start from. In the chat view, these messages are marked "The agents may start from this message". That means your messages, the agent's messages, every command it ran and what it printed, and the contents of every file it read or wrote in that part, with the sensitive details in part 2 deleted. It is uploaded to our server only to create the simulated you and run the three replays.\par
If you came from Prolific, we also keep your Prolific ID with what you upload, so that we can pay you. If you withdraw, we delete your chats, the replays and your answers, but keep your Prolific ID and which completion code you were given, so that you can still be paid and can't take part twice. We delete your Prolific ID 2 years after the study ends.\par
\paragraph{2. Sensitive details deleted before uploading}
Before anything is uploaded, our website deletes these sensitive details and puts a placeholder in their place:\par
text shaped like a key, token or password, with a note saying what it was, such as "[AWS access key removed by the study]"\par
your user name with "participant"\par
email addresses with person1@example.com, person2@example.com and so on\par
computer names with host1, host2 and so on, and network addresses with 192.0.2.1, 192.0.2.2 and so on\par
the folders between your home folder and your project folder with folder1, folder2 and so on\par
account names in GitHub and GitLab addresses with account1, account2 and so on\par
names in git author lines with Person 1, Person 2 and so on\par
Each detail gets the same placeholder everywhere, so the agents' copy of your files still matches the chat. Our server deletes them again when the chats arrive.\par
\paragraph{3. What the check can miss}
The deletion only finds text that matches fixed patterns. A chat can still contain a password, a name, your project's name, or code you're not allowed to share. You decide what to upload: on the next page, press Read what would be uploaded to read all of a chat before you select it.\par
\paragraph{4. Who reads it}
Our server sends the chats to OpenAI and to Amazon Web Services, which runs Anthropic's Claude models. Both process them in the United States.\par
An OpenAI model reads whether you pushed back on or corrected the agent at each marked message. We prefer a message where you did, and otherwise use one of the marked messages. The models then create the simulated you from that chat.\par
Three coding agents then continue that chat from that message. Each works on a copy of the files the chat showed, in its own isolated container that can reach only its model provider.\par
OpenAI doesn't use this data to train its models, and keeps it for up to 30 days to detect abuse. Amazon Web Services doesn't store it and doesn't share it with Anthropic.\par
The researchers running the study read what we keep (part 5) and your answers.\par
\paragraph{5. How long we keep it}
Once the three replays are ready, we delete everything you uploaded and everything made from it, including the simulated you. We keep, for 2 years after the study ends, your first message, a short summary of what happened before the message where the agents start, that message, a short summary of what you wanted from the replayed part, and the three replays (what the agents and the simulated you wrote, and what the agents did in the copy of your project), and your answers. Then we delete them. If our server stops with an error, we keep what you sent until we have looked at the error, then delete it. We use them only for this research. Apart from the model providers in part 4, we don't share them with anyone outside the research team, and we never publish or release them as a dataset. We publish only the study's results: figures, and short excerpts with nothing that identifies you. Before publishing an excerpt, we remove names, user names, email addresses and file paths from it. We delete them sooner whenever you ask.\par
\paragraph{6. Your rights}
Taking part is voluntary. You can stop at any time without giving a reason and without any disadvantage. Until you finish the study, you can withdraw with the Withdraw and delete my data button at the top right of every page after you upload your chats. After you finish, contact the researchers ([researcher contact]) to have your data deleted.\par
If you withdraw before finishing, we pay you \pounds 3 as a bonus through Prolific. If you ask us to delete your data after finishing, you keep your \pounds 11.\par
Withdrawing deletes from our server everything you uploaded, the three replays and your answers. You can also contact the researchers to have them deleted.\par
Copies the model providers already received are kept as part 4 says.\par
If you withdraw while the agents are running (up to about 25 minutes of the wait), your withdrawal counts at once and we delete your data as soon as they end.\par
If you came from Prolific and stop before finishing without withdrawing, we don't pay you, with two exceptions. If the eligibility check finds no usable chat, we pay you \pounds 0.50 for the check. If our server finds after you upload your chats that none can be used, we pay you \pounds 3 as a bonus.\par
For questions about your rights as a participant, contact the [institutional review board] office ([ethics office contact]). The [institutional review board] approved this study ([approval reference]).\par
\paragraph{7. Risks and benefits}
The main risk is that something private in your chat is uploaded to our server and passed to the model providers while the replays run. Three things reduce it: sensitive details are deleted before uploading, you choose which chats to upload after reading them, and we delete your chats once the replays are ready. What we keep (part 5) can still include text from your chat, such as your first message and what the agents read in the copy of your files. You receive the payment above, and you see how three different coding agents handle a moment from your own work.\par
Continue to choose chats Select the box above to continue.\par
I'm 18 or older, I've read the seven parts above, and I agree.\par

\section{Code, Data, and Compute}
\label{app:reproducibility}

The agents ran in Claude Code 2.1.220 in the adaptation test and the user study, in Codex 0.154.0 in the user study, and in opencode 1.18.29 in the evaluation of coding agents.
In total, the experiments in this paper cost about \$3,700 at API prices.
The models were accessed through hosted APIs, and their parameter counts and the hardware used to serve them are not publicly available.
Building one intent model cost about \$3 on average, and each turn of the user simulator cost less than \$0.01.
The tasks can be rebuilt from the public releases of SWE-Together and SWE-chat with the session identifiers in Table~\ref{tab:task-ids}.

\section{Prompts}
\label{app:prompts}

This appendix gives the prompts of the LLMs and agents in \framework{} and in our experiments.
Parts that change with each task or turn are shown in monospace within braces.
Text that we cut is marked with ``[...]'', and a note in brackets says what each longer cut held.

\setcounter{dbltopnumber}{3}
\renewcommand{\dbltopfraction}{0.95}
\renewcommand{\dblfloatpagefraction}{0.6}
\makeatletter\setlength{\@dblfptop}{0pt}\setlength{\@dblfpbot}{0pt plus 1fil}\makeatother

\begin{figure*}[tp]
\begin{tcolorbox}[colback=gray!5!white, colframe=gray!75!black, title=Drafting agent, fontupper=\small]
\noindent You are building the intent model of one recorded session between a user and a coding agent. A simulated user and an assessor read the intent model while a different agent works on the same task in a new run. The intent model also records where, in this recorded session, each criterion changed and each condition was met.\par
\smallskip
\noindent Everything you need is in this folder. Read nothing outside it and fetch nothing from the web.\par
\hangindent=2em\hangafter=1\noindent\hspace*{1em}\makebox[1em][l]{\textbullet}\texttt{\{rules\}}\allowbreak{}: the rules. They set out a process of eight steps and the rules each step applies.\par
\hangindent=2em\hangafter=1\noindent\hspace*{1em}\makebox[1em][l]{\textbullet}\texttt{\{format\}}\allowbreak{}: the exact JSON format of the intent model, key by key.\par
\hangindent=2em\hangafter=1\noindent\hspace*{1em}\makebox[1em][l]{\textbullet}\texttt{\{starting\_environment\}}\allowbreak{}: what every run of this task starts in.\par
\hangindent=2em\hangafter=1\noindent\hspace*{1em}\makebox[1em][l]{\textbullet}\texttt{\{starting\_workspace\}}\allowbreak{}: every file and git ref of the workspace a run starts in.\par
\hangindent=2em\hangafter=1\noindent\hspace*{1em}\makebox[1em][l]{\textbullet}\texttt{\{recorded\_session\}}\allowbreak{}: the recorded session as a run shows it, with every user turn listed first.\par
\smallskip
\noindent You write two files: the intent model, and a record of your decisions.\par
\smallskip
\noindent What to do\par
\hangindent=2.4em\hangafter=1\noindent\hspace*{1em}\makebox[1.4em][l]{1.}Read the rules whole, then the format, then the starting environment.\par
\hangindent=2.4em\hangafter=1\noindent\hspace*{1em}\makebox[1.4em][l]{2.}Read every line of the recorded session, to its last line. Decide nothing before you have read to the end.\par
\hangindent=2.4em\hangafter=1\noindent\hspace*{1em}\makebox[1.4em][l]{3.}Walk the process in the rules, steps 1 to 8, in order. At each step, make that step's decisions under that step's rules and write them into the intent model. \textit{[\dots{} how to go back when a later step shows an earlier decision was wrong \dots{}]}\par
\hangindent=2.4em\hangafter=1\noindent\hspace*{1em}\makebox[1.4em][l]{4.}At step 7, write a table with one row for every change the intent model records, and correct the intent model where a row of the table does not hold.\par
\hangindent=2.4em\hangafter=1\noindent\hspace*{1em}\makebox[1.4em][l]{5.}At step 8, trace both replays the rules name, intent by intent, and write each in your record.\par
\hangindent=2.4em\hangafter=1\noindent\hspace*{1em}\makebox[1.4em][l]{6.}Finish your record. \textit{[\dots{} its six parts \dots{}]}\par
\end{tcolorbox}
\caption{Prompt for the drafting agent, which drafts the intent model.}
\label{fig:prompt-draft}
\end{figure*}

\begin{figure*}[tp]
\begin{tcolorbox}[colback=gray!5!white, colframe=gray!75!black, title=Reviewing agent, fontupper=\small]
\noindent You are checking one intent model against the recorded session it was built from. Do not edit or write any file. You may also run \texttt{\{probe\_command\}}\allowbreak{}, which runs a command in a new container of the environment every run starts in.\par
\smallskip
\hangindent=2em\hangafter=1\noindent\hspace*{1em}\makebox[1em][l]{\textbullet}\texttt{\{rules\}}\allowbreak{}, \texttt{\{starting\_environment\}}\allowbreak{}, \texttt{\{starting\_workspace\}}\allowbreak{} and \texttt{\{recorded\_session\}}\allowbreak{}: as in the drafting agent's prompt\par
\hangindent=2em\hangafter=1\noindent\hspace*{1em}\makebox[1em][l]{\textbullet}\texttt{\{intent\_model\}}\allowbreak{}: the intent model, printed compactly.\par
\smallskip
\noindent What to do\par
\hangindent=2.4em\hangafter=1\noindent\hspace*{1em}\makebox[1.4em][l]{1.}Read the rules whole, then the starting environment, then every line of the recorded session, then the intent model.\par
\hangindent=2.4em\hangafter=1\noindent\hspace*{1em}\makebox[1.4em][l]{2.}Walk the process in the rules, steps 1 to 8, in order. At each step, take every decision the intent model made at that step and check it against each rule of that step. \textit{[\dots{} what each of the eight steps checks \dots{}]}\par
\hangindent=2.4em\hangafter=1\noindent\hspace*{1em}\makebox[1.4em][l]{3.}Report each defect you can show with events: the rule it breaks, the items, and the events. Leave out anything you cannot show. The intent model's own reasons say what its author meant. They are claims, not evidence.\par
\smallskip
\noindent Reply with exactly this and nothing else, one line per defect, in step order:\par
\noindent FINDINGS \texttt{\{task\_name\}}\allowbreak{}\par
\hangindent=2em\hangafter=1\noindent\hspace*{1em}\makebox[1em][l]{\textbullet}rule N | \texttt{\{item\_ids\_or\_turn\}}\allowbreak{} | events [x, y] | \texttt{\{what\_is\_wrong\_and\_what\_would\_fix\_it\_in\_one\_sentence\}}\allowbreak{}\par
\noindent AUDIT \texttt{\{task\_name\}}\allowbreak{} | \texttt{\{number\_of\_defect\_lines\}}\allowbreak{}\par
\end{tcolorbox}
\caption{Prompt for the reviewing agent, which checks the draft of the intent model for issues.}
\label{fig:prompt-review}
\end{figure*}

\clearpage

\begin{figure*}[tp]
\begin{tcolorbox}[colback=gray!5!white, colframe=gray!75!black, title=Revising agent, fontupper=\small]
\noindent You are repairing one intent model. Another agent built it from the recorded session, and a checker then walked the rules against it and reported defects. Your job is to decide each reported defect, repair the ones that hold, and make sure no repair breaks another rule.\par
\smallskip
\hangindent=2em\hangafter=1\noindent\hspace*{1em}\makebox[1em][l]{\textbullet}\texttt{\{rules\}}\allowbreak{}, \texttt{\{format\}}\allowbreak{}, \texttt{\{starting\_environment\}}\allowbreak{}, \texttt{\{starting\_workspace\}}\allowbreak{} and \texttt{\{recorded\_session\}}\allowbreak{}: as in the drafting agent's prompt\par
\hangindent=2em\hangafter=1\noindent\hspace*{1em}\makebox[1em][l]{\textbullet}\texttt{\{intent\_model\}}\allowbreak{}: the intent model. This is the file you edit.\par
\hangindent=2em\hangafter=1\noindent\hspace*{1em}\makebox[1em][l]{\textbullet}\texttt{\{findings\}}\allowbreak{}: the checker's report, one line per defect.\par
\smallskip
\noindent What to do\par
\hangindent=2.4em\hangafter=1\noindent\hspace*{1em}\makebox[1.4em][l]{1.}Read the rules whole, then the format, then the starting environment, then every line of the recorded session, then the intent model and the findings.\par
\hangindent=2.4em\hangafter=1\noindent\hspace*{1em}\makebox[1.4em][l]{2.}For each finding: read the rule it names, the items and the events. Decide whether the rule is broken there. Where it is, repair it, changing only what the repair needs. Where it is not, leave the items.\par
\hangindent=2.4em\hangafter=1\noindent\hspace*{1em}\makebox[1.4em][l]{3.}Re-check every item you changed, added or removed against every rule of its step and of every later step. For every intent you changed, and every intent waiting on it, trace again both replays the rules name. \textit{[\dots{} which step each kind of item belongs to \dots{}]}\par
\hangindent=2.4em\hangafter=1\noindent\hspace*{1em}\makebox[1.4em][l]{4.}Write your record: for each finding, whether it holds, why, with events, and what you changed. \textit{[\dots{} three further parts \dots{}]}\par
\end{tcolorbox}
\caption{Prompt for the revising agent, which revises the draft of the intent model based on the review.}
\label{fig:prompt-revise}
\end{figure*}

\begin{figure*}[tp]
\begin{tcolorbox}[colback=gray!5!white, colframe=gray!75!black, title=Rules, fontupper=\small]
\noindent An intent is one outcome the user would recognise as their own checkpoint. A criterion is one thing that must hold for an intent to be satisfied. A condition says when an intent may be raised or stops being wanted (is dropped).\par
\smallskip
\noindent The intent model's words are read by a simulated user and an assessor while a different agent works on the task its own way. The record of changes describes this recorded session.\par
\smallskip
\noindent Step 1. The user's turns\par
\hangindent=2.4em\hangafter=1\noindent\hspace*{1em}\makebox[1.4em][l]{1.}Label each user turn as an ask, constraint, correction, command, report, acceptance, answer, control or noise. A turn expresses every intent its content is about.\par
\hangindent=2.4em\hangafter=1\noindent\hspace*{1em}\makebox[1.4em][l]{2.}A turn that only keeps the exchange going expresses nothing.\par
\noindent \textit{[\dots{} 6 further rules: tool choices, pasted text, content a replay cannot produce, tasks that start from a later user message \dots{}]}\par
\noindent Step 2. Dividing the turns into intents\par
\hangindent=2.4em\hangafter=1\noindent\hspace*{1em}\makebox[1.4em][l]{9.}One outcome the user would recognise as their own checkpoint is one intent.\par
\noindent \textit{[\dots{} 5 further rules: repository actions, questions, replaced requests, citing turns \dots{}]}\par
\noindent Step 3. Criteria\par
\hangindent=2.4em\hangafter=1\noindent\hspace*{1em}\makebox[1.4em][l]{15.}A criterion is explicit where the user stated it, and revealed where a delivery and the user's acceptance of it show what a broad request needed.\par
\noindent \textit{[\dots{} 5 further rules: what to reach and what not to do, the scope of a criterion, how much evidence it needs \dots{}]}\par
\noindent Step 4. Names and wording\par
\hangindent=2.4em\hangafter=1\noindent\hspace*{1em}\makebox[1.4em][l]{21.}The intent model names only what the user named, what the opening request gives, or what the starting environment holds, and never what the recorded agent found, chose or produced.\par
\noindent \textit{[\dots{} 2 further rules \dots{}]}\par
\noindent Step 5. Evidence\par
\hangindent=2.4em\hangafter=1\noindent\hspace*{1em}\makebox[1.4em][l]{24.}An item rests on what a run's conversation shows.\par
\noindent \textit{[\dots{} 1 further rule \dots{}]}\par
\noindent Step 6. Conditions\par
\hangindent=2.4em\hangafter=1\noindent\hspace*{1em}\makebox[1.4em][l]{26.}A raising condition must pass this question: could an agent that does everything the user wanted, in a reasonable way, fail to reach it?\par
\noindent \textit{[\dots{} 4 further rules \dots{}]}\par
\noindent Step 7. The record of changes\par
\hangindent=2.4em\hangafter=1\noindent\hspace*{1em}\makebox[1.4em][l]{31.}Each change sits at the first agent turn whose shown text shows it.\par
\noindent \textit{[\dots{} 2 further rules \dots{}]}\par
\noindent Step 8. The environment and reaching an end\par
\hangindent=2.4em\hangafter=1\noindent\hspace*{1em}\makebox[1.4em][l]{34.}An item no run in the starting environment can meet is restated as the outcome the environment can check, or left out.\par
\hangindent=2.4em\hangafter=1\noindent\hspace*{1em}\makebox[1.4em][l]{35.}Every intent can be raised by the simulated user and end, satisfied or dropped, in two replays: one where the agent gets everything right the first time, and one where it works another reasonable way.\par
\end{tcolorbox}
\caption{Rules that the drafting, reviewing and revising agents read.}
\label{fig:prompt-rules}
\end{figure*}

\clearpage

\begin{figure*}[tp]
\begin{tcolorbox}[colback=gray!5!white, colframe=gray!75!black, title=Assess, fontupper=\small]
\noindent \textbf{[System prompt]}\par
\noindent You play one specific software user assessing a coding agent's latest block. Judge only from the visible agent block. Do not request more data or write a user message.\par
\smallskip
\noindent Criterion results are changes from the shown prior state, not current-state labels:\par
\hangindent=2em\hangafter=1\noindent\hspace*{1em}\makebox[1em][l]{\textbullet}prior unsatisfied and still unmet $\rightarrow$ no\_change;\par
\hangindent=2em\hangafter=1\noindent\hspace*{1em}\makebox[1em][l]{\textbullet}prior unsatisfied and newly met $\rightarrow$ satisfied;\par
\hangindent=2em\hangafter=1\noindent\hspace*{1em}\makebox[1em][l]{\textbullet}prior satisfied and still met $\rightarrow$ no\_change;\par
\hangindent=2em\hangafter=1\noindent\hspace*{1em}\makebox[1em][l]{\textbullet}prior satisfied and newly unmet $\rightarrow$ dissatisfied.\par
\smallskip
\noindent \textit{[\dots{} how to judge a criterion that names what to verify \dots{}]}\par
\smallskip
\noindent For the lifecycle conditions you are asked to judge, use met or not\_met. Conditions are latched triggers: a condition whose prior state is met remains met. For a prior not\_met condition, return met only when affirmative evidence shows the entire described trigger is now achieved; partial or related progress remains not\_met.\par
\smallskip
\noindent Return exactly one row for every REQUIRED item and no other item. Return only this JSON object shape:\par
\noindent {\raggedright\texttt{\{"criterion\_assessments": [\{"criterion\_id": "...", "result": "...",}\par}\par
\noindent {\raggedright\texttt{"reason": "concise audit rationale", "evidence\_refs": [], ...\}],}\par}\par
\noindent {\raggedright\texttt{"condition\_assessments": [ ... the same fields ... ]\}}\par}\par
\smallskip
\noindent \textit{[\dots{} rules for evidence references \dots{}]}\par
\smallskip
\noindent \raggedright\textbf{[User message]}\par
\noindent Current private rubric:\par
\noindent Intent \texttt{\{intent\_id\}}\allowbreak{} [\texttt{\{intent\_engagement\}}\allowbreak{}, \texttt{\{intent\_satisfaction\}}\allowbreak{}; REQUIRED]: \texttt{\{intent\_description\}}\allowbreak{}\par
\hangindent=3em\hangafter=1\noindent\hspace*{2em}\makebox[1em][l]{\textbullet}\texttt{\{criterion\_id\}}\allowbreak{} [prior=\texttt{\{criterion\_status\}}\allowbreak{}]: \texttt{\{criterion\_description\}}\allowbreak{}\par
\hangindent=3em\hangafter=1\noindent\hspace*{3em}verify (conversation): \texttt{\{verification\_target\_description\}}\allowbreak{}\par
\noindent \textit{[\dots{} one such entry for every raised or dropped intent and each of its criteria \dots{}]}\par
\noindent Conditions written in text (REQUIRED; retain the current state when unchanged):\par
\hangindent=3em\hangafter=1\noindent\hspace*{2em}\makebox[1em][l]{\textbullet}\texttt{\{condition\_id\}}\allowbreak{} [prior=\texttt{\{condition\_status\}}\allowbreak{}]: \texttt{\{condition\_description\}}\allowbreak{}\par
\smallskip
\noindent Prior conversation:\par
\noindent \texttt{\{prior\_conversation\}}\allowbreak{}\par
\smallskip
\noindent Latest visible agent block:\par
\noindent \texttt{\{latest\_agent\_turn\}}\allowbreak{}\par
\smallskip
\noindent Assess the complete REQUIRED population and no other items.\par
\end{tcolorbox}
\caption{Prompt for the Assess step of the user simulator (GPT-5.6 Luna).
The prompt calls an agent turn a block.}
\label{fig:prompt-assess}
\end{figure*}

\begin{figure*}[tp]
\begin{tcolorbox}[colback=gray!5!white, colframe=gray!75!black, title=Respond, fontupper=\small]
\noindent \textbf{[System prompt]}\par
\noindent You play one specific software user deciding and writing the next substantive message to a coding agent. The supplied reduced state is fact; do not reassess it. Decide what this user naturally needs to communicate now, write that message, and map the message's substantive clauses in the same response.\par
\smallskip
\noindent The interaction profile is how you write; it does not decide what you say.\par
\smallskip
\noindent Work in this order.\par
\smallskip
\hangindent=2.4em\hangafter=1\noindent\hspace*{1em}\makebox[1.4em][l]{1.}Decide what this message says by the rules below: which want it raises, what it asks for, or what it answers. Write that in draft, in the plainest words.\par
\hangindent=2.4em\hangafter=1\noindent\hspace*{1em}\makebox[1.4em][l]{2.}Write message: what the draft says, typed the way this person types. Apply every habit in the writing style to it. The message asks for exactly what the draft asks for: rewording never adds a request, drops one, swaps it for another, or changes which want is raised.\par
\smallskip
\noindent Choose the smallest natural next turn. The latest visible agent block defines the immediate conversational situation; the reduced state lists what this user wants in no particular order, and it is not a checklist to bundle into one message.\par
\smallskip
\noindent Write it as briefly as this user would. \textit{[\dots{} what a brief message leaves out \dots{}]}\par
\smallskip
\noindent In the reduced state, an intent marked pursuing is one this user has already put into words in this conversation. An intent marked available has not been raised yet, and raising one is what advancing means.\par
\noindent \textit{[\dots{} when to raise an intent again, intents that the latest agent turn changed, and intents to find something out \dots{}]}\par
\smallskip
\noindent Apply this priority:\par
\hangindent=2.4em\hangafter=1\noindent\hspace*{1em}\makebox[1.4em][l]{1.}The agent's question, approval request, or menu is an invitation, not an obligation. \textit{[\dots{} when to answer it and what to say instead \dots{}]}\par
\hangindent=2.4em\hangafter=1\noindent\hspace*{1em}\makebox[1.4em][l]{2.}When the latest agent block is unfinished, the work is under way, and your message is about that work. \textit{[\dots{} checking the step against the limits the state lists \dots{}]}\par
\hangindent=2.4em\hangafter=1\noindent\hspace*{1em}\makebox[1.4em][l]{3.}When the agent reports completing the current work, start from your own previous message. \textit{[\dots{} pressing a doubt, or advancing to one natural next intent \dots{}]}\par
\hangindent=2.4em\hangafter=1\noindent\hspace*{1em}\makebox[1.4em][l]{4.}Use a bare acknowledgment only when no next planned ask is called for.\par
\smallskip
\noindent \textit{[\dots{} rules on menus, bundling, and which intents and criteria the message expresses \dots{}]}\par
\smallskip
\noindent Return only this JSON object:\par
\noindent {\raggedright\texttt{\{"draft": "...",}\par}\par
\noindent {\raggedright\texttt{"message": "...",}\par}\par
\noindent {\raggedright\texttt{"clause\_grounding": [\{"clause": "...", "intent\_ids": [], ...\}],}\par}\par
\noindent {\raggedright\texttt{"expressed\_intent\_ids": [], "expressed\_criterion\_ids": [],}\par}\par
\noindent {\raggedright\texttt{\# ... further fields: withdrawn intents, confidence, reasons for each choice}\par}\par
\noindent {\raggedright\texttt{\}}\par}\par
\smallskip
\noindent \raggedright\textbf{[User message]}\par
\noindent Writing style: \texttt{\{style\_profile\}}\allowbreak{}\par
\smallskip
\noindent Reduced state:\par
\noindent Intent \texttt{\{intent\_id\}}\allowbreak{} [\texttt{\{intent\_engagement\}}\allowbreak{}, \texttt{\{intent\_satisfaction\}}\allowbreak{}]\texttt{\{already\_asked\_note\}}\allowbreak{}\texttt{\{moved\_by\_latest\_block\_note\}}\allowbreak{}: \texttt{\{intent\_description\}}\allowbreak{}\par
\hangindent=3em\hangafter=1\noindent\hspace*{2em}\makebox[1em][l]{\textbullet}\texttt{\{criterion\_id\}}\allowbreak{} [\texttt{\{criterion\_status\}}\allowbreak{}]: \texttt{\{criterion\_description\}}\allowbreak{}\par
\hangindent=2em\hangafter=1\noindent\hspace*{2em}still unmet because: \texttt{\{assess\_reason\}}\allowbreak{}\par
\noindent \textit{[\dots{} one such entry for every intent that is not blocked \dots{}]}\par
\smallskip
\noindent Prior conversation:\par
\noindent \texttt{\{prior\_conversation\}}\allowbreak{}\par
\smallskip
\noindent Latest visible agent block:\par
\noindent \texttt{\{latest\_agent\_turn\}}\allowbreak{}\par
\smallskip
\noindent Decide and render the required user message.\par
\end{tcolorbox}
\caption{Prompt for the Respond step of the user simulator (GPT-5.6 Luna).
The prompt calls the style profile the interaction profile, an intent a want, an agent turn a block, and the intent state the reduced state, in which an open intent is available and a raised intent is pursuing.}
\label{fig:prompt-respond}
\end{figure*}

\clearpage

\begin{figure*}[tp]
\begin{tcolorbox}[colback=gray!5!white, colframe=gray!75!black, title=SWE-Interact user simulator, fontupper=\small]
\noindent \textbf{[System prompt]}\par
\noindent You are role-playing as a user talking to a coding assistant. Stay in character at all times. Never break character or mention that you are an AI.\par
\smallskip
\noindent You cannot inspect the repository. You know only what the conversation tells you.\par
\smallskip
\noindent \textit{[\dots{} a note on the source and license of the rules below \dots{}]}\par
\smallskip
\noindent \textbf{Shared invariants}\par
\smallskip
\noindent You are emulating a working developer talking to a coding assistant. Respond like you would talk to a teammate on Slack. Everything below is private behavior.\par
\smallskip
\noindent \textit{[\dots{} rules on what not to mention, what you know, precedence, and conversation \dots{}]}\par
\smallskip
\noindent \textbf{Interaction style: busy colleague}\par
\smallskip
\noindent You are busy and expect the agent to do real repository work. Keep replies short, natural, direct, and a little offhand. Do not conduct long clarification interviews. For mechanics outside the actual goal, make it the agent's call.\par
\smallskip
\noindent \textbf{Disclosure pattern: vibecoder}\par
\smallskip
\noindent Open with a brief, vague overview. Let the agent explore and propose a quick plan without tightening it into the complete task. Once the agent shows an implementation, reveal missing requirements slowly as the implementation makes them relevant.\par
\smallskip
\hangindent=2em\hangafter=1\noindent\hspace*{1em}\makebox[1em][l]{\textbullet}Reveal one concrete requested change at a time, not a checklist or broad category.\par
\noindent \textit{[\dots{} 6 further rules on what to reveal and when to approve \dots{}]}\par
\hangindent=2em\hangafter=1\noindent\hspace*{1em}\makebox[1em][l]{\textbullet}Approve only after every externally visible part of the task block has been addressed.\par
\smallskip
\noindent \textbf{Private task block}\par
\smallskip
\noindent The complete goal consists of the following reviewed outcomes.\par
\smallskip
\hangindent=2em\hangafter=1\noindent\hspace*{1em}\makebox[1em][l]{\textbullet}\texttt{\{intent\_description\}}\allowbreak{}\par
\hangindent=3em\hangafter=1\noindent\hspace*{2em}\makebox[1em][l]{\textbullet}Done when: \texttt{\{criterion\_description\}}\allowbreak{}\par
\hangindent=1em\hangafter=1\noindent\hspace*{1em}\textit{[\dots{} one line for each criterion and one entry for each intent \dots{}]}\par
\smallskip
\noindent The user speaks at this point; that decision is already made and is not yours. Write the message they send.\par
\smallskip
\noindent Reply as JSON: \texttt{\{"message": "<the next user message>"\}}\par
\smallskip
\noindent \raggedright\textbf{[User message]}\par
\noindent {\raggedright\texttt{\{"conversation": [\{"content": \texttt{\{message\_text\}}\allowbreak{}, "role": \texttt{\{user\_or\_assistant\}}\allowbreak{}\}, ...]\}}\par}\par
\end{tcolorbox}
\caption{Prompt for the SWE-Interact user simulator in our comparison (GPT-5.6 Luna).
It consists of the rules of SWE-Interact's released prompt, without the passages that claim access to the repository, followed by the intents and criteria of our intent model.}
\label{fig:prompt-swe-interact}
\end{figure*}

\begin{figure*}[tp]
\begin{tcolorbox}[colback=gray!5!white, colframe=gray!75!black, title=Goal-prompted simulator, fontupper=\small]
\noindent \textbf{[System prompt]}\par
\noindent You are role-playing as the user in a conversation with a coding assistant about your own repository. Stay in character: never say you are an AI or that you are playing a part.\par
\smallskip
\noindent You are given the conversation so far. Write the single next message you would send. Do not invent facts about the repository or the tests that the conversation has not told you.\par
\smallskip
\noindent What you want from this work:\par
\smallskip
\noindent \textbf{Private task block}\par
\smallskip
\noindent The complete goal consists of the following reviewed outcomes. Their order below is lexicographic and carries no historical disclosure or priority information.\par
\smallskip
\hangindent=2em\hangafter=1\noindent\hspace*{1em}\makebox[1em][l]{\textbullet}\texttt{\{intent\_description\}}\allowbreak{}\par
\hangindent=3em\hangafter=1\noindent\hspace*{2em}\makebox[1em][l]{\textbullet}Done when: \texttt{\{criterion\_description\}}\allowbreak{}\par
\hangindent=1em\hangafter=1\noindent\hspace*{1em}\textit{[\dots{} one line for each criterion and one entry for each intent \dots{}]}\par
\smallskip
\noindent The user speaks at this point; that decision is already made and is not yours. Do not stay silent and do not return an empty message. Write the message they send.\par
\smallskip
\noindent Reply as JSON: \texttt{\{"message": "<the next user message>"\}}\par
\smallskip
\noindent \raggedright\textbf{[User message]}\par
\noindent {\raggedright\texttt{\{"conversation": [\{"content": \texttt{\{message\_text\}}\allowbreak{}, "role": \texttt{\{user\_or\_assistant\}}\allowbreak{}\}, ...]\}}\par}\par
\end{tcolorbox}
\caption{Prompt for the goal-prompted simulator (GPT-5.6 Luna).}
\label{fig:prompt-simple-user}
\end{figure*}

\clearpage

\begin{figure*}[tp]
\begin{tcolorbox}[colback=gray!5!white, colframe=gray!75!black, title=Judging reply similarity (single-turn), fontupper=\small]
\noindent \textbf{[System prompt]}\par
\noindent You are comparing two replies a developer could send to a coding assistant at the same point in a conversation: the reply the real developer sent, and a candidate reply. You are not told where the candidate came from.\par
\smallskip
\noindent The input is JSON: \texttt{recorded\_reply} is the reply the real developer sent, \texttt{candidate} is the candidate reply, and \texttt{assistant\_last\_message} is the assistant's last message before the developer replied, its prose only.\par
\smallskip
\noindent Use \texttt{assistant\_last\_message} for one thing only: to tell what a reply points at. Every judgment compares the candidate with the real developer's reply.\par
\smallskip
\noindent Judge two things, separately: \textbf{intent} (what the reply asks the assistant to do) and \textbf{style} (how the reply is written). For each, first list every difference you find, then give the score the rubric sets for those differences.\par
\smallskip
\noindent For each difference between the two replies, ask: \textbf{would the assistant do something different because of it?}\par
\hangindent=2em\hangafter=1\noindent\hspace*{1em}\makebox[1em][l]{\textbullet}If yes, it is an intent difference.\par
\hangindent=2em\hangafter=1\noindent\hspace*{1em}\makebox[1em][l]{\textbullet}If no, it is a style difference.\par
\noindent \textit{[\dots{} worked examples of the rule \dots{}]}\par
\smallskip
\noindent \textit{[\dots{} 9 rules for reading both replies, such as "A request in question form is an order" \dots{}]}\par
\smallskip
\noindent \textbf{1. intent}\par
\smallskip
\noindent List every intent difference, each with one of these types:\par
\noindent \textit{[\dots{} a table that defines the 10 types \dots{}]}\par
\smallskip
\noindent \textbf{Intent score.} No differences gives 5. Otherwise, the score is the lowest level any listed difference reaches:\par
\smallskip
\noindent
\hspace*{1em}\begin{tabular}{@{}rl@{}}
score & reached by \\ \hline
5 & no intent difference \\
4 & \texttt{fact\_left\_out}, \texttt{fact\_added} \\
3 & \texttt{different\_kind\_of\_turn}, \texttt{request\_missing}, \texttt{request\_added} \\
2 & \texttt{different\_step}, \texttt{different\_outcome} \\
1 & \texttt{other\_work}, \texttt{nothing\_where\_asked}, \texttt{request\_where\_none} \\
\end{tabular}
\par
\smallskip
\noindent \textit{[\dots{} worked cases \dots{}]}\par
\smallskip
\noindent \textbf{2. style}\par
\smallskip
\noindent Judge only how the reply is written, as if both asked for the same thing.\par
\smallskip
\noindent Judge the five dimensions below one at a time. For each, first say in a few words what the developer's reply does, then what the candidate does, then mark it:\par
\hangindent=2em\hangafter=1\noindent\hspace*{1em}\makebox[1em][l]{\textbullet}0: the candidate does the same as the developer's reply;\par
\hangindent=2em\hangafter=1\noindent\hspace*{1em}\makebox[1em][l]{\textbullet}1: somewhat different;\par
\hangindent=2em\hangafter=1\noindent\hspace*{1em}\makebox[1em][l]{\textbullet}2: clearly different.\par
\smallskip
\noindent
\hspace*{1em}\begin{tabular}{@{}ll@{}}
dimension & what it covers \\ \hline
\texttt{length\_and\_detail} & how much the reply says \textit{[\dots{}]} \\
\texttt{tone} & how the reply sounds \textit{[\dots{}]} \\
\texttt{sentence\_form} & how the reply is built \textit{[\dots{}]} \\
\texttt{reference} & how the reply names things \textit{[\dots{}]} \\
\texttt{typing\_habits} & how the reply is typed \textit{[\dots{}]} \\
\end{tabular}
\par
\smallskip
\noindent \textbf{Style score.} Add the five marks (0 to 10):\par
\smallskip
\noindent
\hspace*{1em}\begin{tabular}{@{}rl@{}}
score & total of the marks \\ \hline
5 & 0 \\
4 & 1 or 2 \\
3 & 3 or 4 \\
2 & 5 or 6 \\
1 & 7 or more \\
\end{tabular}
\par
\smallskip
\noindent \textit{[\dots{} worked cases \dots{}]}\par
\smallskip
\noindent \textbf{What to write}\par
\smallskip
\noindent Answer in the JSON shape you are given: \texttt{intent\_differences}, \texttt{intent}, \texttt{intent\_reason}, \texttt{style\_dimensions}, \texttt{style}, \texttt{style\_reason}. \textit{[\dots{} what each field holds \dots{}]}\par
\smallskip
\noindent \raggedright\textbf{[User message]}\par
\noindent {\raggedright\texttt{\{"assistant\_last\_message": \texttt{\{assistant\_last\_message\}}\allowbreak{}, "candidate": \texttt{\{candidate\_reply\}}\allowbreak{}, "recorded\_reply": \texttt{\{recorded\_reply\}}\allowbreak{}\}}\par}\par
\end{tcolorbox}
\caption{Prompt to judge single-turn reply similarity (GPT-6.1 Sol).}
\label{fig:prompt-judge-reply}
\end{figure*}

\begin{figure*}[tp]
\begin{tcolorbox}[colback=gray!5!white, colframe=gray!75!black, title=Judging reply similarity (multi-turn), fontupper=\small]
\noindent \textbf{[System prompt]}\par
\noindent You are comparing two stretches of replies a developer could send to a coding assistant over the same few consecutive turns of a conversation: the replies the real developer sent, and a candidate's replies. You are not told where the candidate came from.\par
\smallskip
\noindent \textit{[\dots{} the input format, and the use of the agent messages as in the single-turn prompt \dots{}]}\par
\smallskip
\noindent \textbf{Reading a stretch}\par
\smallskip
\noindent Judge the stretch as a whole, not turn by turn.\par
\smallskip
\noindent Work in this order:\par
\hangindent=2.4em\hangafter=1\noindent\hspace*{1em}\makebox[1.4em][l]{1.}List everything the developer asked for over the whole stretch, whatever turn each request came at.\par
\hangindent=2.4em\hangafter=1\noindent\hspace*{1em}\makebox[1.4em][l]{2.}List everything the candidate asked for over the whole stretch, whatever turn each request came at.\par
\hangindent=2.4em\hangafter=1\noindent\hspace*{1em}\makebox[1.4em][l]{3.}Compare the two lists. A request on both lists is the same request, whichever turn each made it at: raising it a turn earlier or later is not a difference. A request on only one list is a difference.\par
\hangindent=2.4em\hangafter=1\noindent\hspace*{1em}\makebox[1.4em][l]{4.}For style, compare how the developer's replies are written, taken together, with how the candidate's are written, taken together.\par
\smallskip
\noindent \textit{[\dots{} the rest is the single-turn prompt from "Judge two things, separately" to its end \dots{}]}\par
\smallskip
\noindent \raggedright\textbf{[User message]}\par
\noindent {\raggedright\texttt{\{"candidate\_replies": [\texttt{\{candidate\_reply\_1\}}\allowbreak{}, \texttt{\{candidate\_reply\_2\}}\allowbreak{}, \texttt{\{candidate\_reply\_3\}}\allowbreak{}], "developer\_stretch": [\{"assistant\_last\_message": \texttt{\{assistant\_last\_message\_1\}}\allowbreak{}, "recorded\_reply": \texttt{\{recorded\_reply\_1\}}\allowbreak{}\}, ...]\}}\par}\par
\end{tcolorbox}
\caption{Prompt to judge multi-turn reply similarity (GPT-6.1 Sol).}
\label{fig:prompt-judge-stretch}
\end{figure*}

\clearpage

\begin{figure*}[tp]
\begin{tcolorbox}[colback=gray!5!white, colframe=gray!75!black, title=Paraphrasing, fontupper=\small]
\noindent \textbf{[System prompt]}\par
\noindent You are rewording things that were said in one exchange between a person and a coding assistant. Each text keeps its meaning and changes its words.\par
\smallskip
\noindent You are given several texts, numbered. Some are the person's, some are the assistant's, and each says which. Reword each one on its own and return them under the same numbers.\par
\smallskip
\noindent What must not change in any of them:\par
\hangindent=2em\hangafter=1\noindent\hspace*{1em}\makebox[1em][l]{\textbullet}Every fact, figure, file name, path, command, identifier and error message.\par
\hangindent=2em\hangafter=1\noindent\hspace*{1em}\makebox[1em][l]{\textbullet}What the text does. A text that asks stays a question; one that instructs stays an instruction; one that reports what happened stays a report.\par
\hangindent=2em\hangafter=1\noindent\hspace*{1em}\makebox[1em][l]{\textbullet}How sure the speaker is, and how specific, in both directions.\par
\hangindent=2em\hangafter=1\noindent\hspace*{1em}\makebox[1em][l]{\textbullet}\textbf{Every word that negates, exactly as it was.}\par
\hangindent=2em\hangafter=1\noindent\hspace*{1em}\makebox[1em][l]{\textbullet}\textbf{What the speaker asserts about the history of the problem.} Do not add "still", "already", "again", "yet" where they did not write one, and do not remove one they did.\par
\hangindent=2em\hangafter=1\noindent\hspace*{1em}\makebox[1em][l]{\textbullet}The language each text is written in, one text at a time.\par
\hangindent=2em\hangafter=1\noindent\hspace*{1em}\makebox[1em][l]{\textbullet}Any span in backticks, \textbf{the backticks included}, any file path, any @-mention of a file, any URL, and any identifier written with parentheses.\par
\noindent \textit{[\dots{} further detail and examples for each of these, and the handling of markers for pasted text \dots{}]}\par
\smallskip
\noindent For the person's own texts there is one more rule, and it matters more than reading well: \textbf{keep the way this person writes.} Their spelling, including their mistakes. Their length. Their bluntness or their politeness. Their lowercase. Their missing punctuation. A person who types four words gets back four or five words, not a sentence.\par
\smallskip
\noindent \textit{[\dots{} examples of good and bad paraphrases \dots{}]}\par
\smallskip
\noindent For the assistant's texts, keep what it says it did and what it found. Its prose may become plainer, but not shorter in content.\par
\smallskip
\noindent Do not add anything, do not drop anything, and do not comment on the texts or on this instruction.\par
\smallskip
\noindent Answer with one block per text and nothing else:\par
\smallskip
\noindent {\raggedright\texttt{=== text 1 ===}\par}\par
\noindent the first reworded text\par
\noindent {\raggedright\texttt{=== text 2 ===}\par}\par
\noindent the second reworded text\par
\smallskip
\noindent \raggedright\textbf{[User message]}\par
\noindent Text 1, said by \texttt{\{speaker\_1\}}\allowbreak{}:\par
\noindent \texttt{\{text\_1\}}\allowbreak{}\par
\smallskip
\noindent Text 2, said by \texttt{\{speaker\_2\}}\allowbreak{}:\par
\noindent \texttt{\{text\_2\}}\allowbreak{}\par
\smallskip
\noindent \dots{}\par
\end{tcolorbox}
\caption{Prompt to paraphrase the user and agent messages of a session for the rewritten session (GPT-5.6 Luna).}
\label{fig:prompt-reword}
\end{figure*}

\clearpage

\begin{figure*}[tp]
\begin{tcolorbox}[colback=gray!5!white, colframe=gray!75!black, title=Merge decision, fontupper=\small]
\noindent \textbf{[System prompt]}\par
\noindent Two things one person said, one after the other, to a coding assistant. Between them the assistant worked and replied. Decide whether the two could have been said as one message at the point of the first.\par
\smallskip
\noindent They could not, if the second one could only have been written after reading the assistant's reply. That happens two ways: it names something the reply supplied, or it reports having done what the reply told them to do. Needing the first thing to be \textit{done} first is not a barrier: the person can say so as a condition or an order of work.\par
\smallskip
\noindent Two independent requests can be one message:\par
\hangindent=1em\hangafter=1\noindent\hspace*{1em}first: add a dark mode toggle to the settings page\par
\hangindent=1em\hangafter=1\noindent\hspace*{1em}second: also make the sidebar collapsible\par
\hangindent=1em\hangafter=1\noindent\hspace*{1em}{\raggedright\texttt{\{"merge": true, "why": "two unrelated changes, neither needs the other"\}}\par}\par
\smallskip
\noindent \textit{[\dots{} four further examples: material the person supplies, a request that waits on the first, and two that cannot merge \dots{}]}\par
\smallskip
\noindent Answer with one JSON object and nothing else.\par
\smallskip
\noindent \raggedright\textbf{[User message]}\par
\noindent The person's first turn:\par
\noindent \texttt{\{earlier\_turn\}}\allowbreak{}\par
\smallskip
\noindent What the assistant replied:\par
\noindent \texttt{\{assistant\_reply\}}\allowbreak{}\par
\smallskip
\noindent The person's second turn:\par
\noindent \texttt{\{later\_turn\}}\allowbreak{}\par
\end{tcolorbox}
\caption{Prompt to decide whether two consecutive user turns could have been sent as one message (GPT-5.6 Luna).}
\label{fig:prompt-merge-decide}
\end{figure*}

\begin{figure*}[tp]
\begin{tcolorbox}[colback=gray!5!white, colframe=gray!75!black, title=Merge joining phrase, fontupper=\small]
\noindent \textbf{[System prompt]}\par
\noindent A reading has decided that these two things one person said could have been said as one message. Your job is to choose the one word or phrase that joins them, and nothing else.\par
\smallskip
\noindent The merged message is the first thing exactly as written, then your joining phrase, then the second thing exactly as written. You may not add a verb, invent a condition, explain what either one is about, swap their order, correct a spelling, or drop a single word.\par
\smallskip
\noindent Choose from these and nothing else:\par
\hangindent=2em\hangafter=1\noindent\hspace*{1em}\makebox[1em][l]{\textbullet}the two do not wait on each other: "Also, " or "and "\par
\hangindent=2em\hangafter=1\noindent\hspace*{1em}\makebox[1em][l]{\textbullet}the second waits on the first being done: "Then after that, " or "and then "\par
\smallskip
\noindent Use the person's own language. Pick the waiting form only where the second act genuinely cannot happen until the first is done. \textit{[\dots{} an example \dots{}]}\par
\smallskip
\noindent Answer with one JSON object and nothing else:\par
\noindent {\raggedright\texttt{\{"connective": "Also, ", "waits": false\}}\par}\par
\smallskip
\noindent \raggedright\textbf{[User message]}\par
\noindent The person's first turn:\par
\noindent \texttt{\{earlier\_turn\}}\allowbreak{}\par
\smallskip
\noindent The person's second turn:\par
\noindent \texttt{\{later\_turn\}}\allowbreak{}\par
\end{tcolorbox}
\caption{Prompt to choose the phrase that joins two consecutive user turns into one turn (GPT-5.6 Luna).}
\label{fig:prompt-merge-write}
\end{figure*}

\clearpage

\begin{figure*}[tp]
\begin{tcolorbox}[colback=gray!5!white, colframe=gray!75!black, title=Writing the goals, fontupper=\small]
\noindent \textbf{[System prompt]}\par
\noindent \textbf{Write the result the person asked for, as a list of clauses}\par
\smallskip
\noindent You read one recorded session between a person and a coding agent, and you write the user target: the result the person asked for over the session, including what their requests necessarily require, and excluding what the recorded agent did on its own.\par
\smallskip
\noindent The target is used to grade a replay, and is never shown to the replayed agent or the simulated user. \textit{[\dots{}]} So every clause says something that can be seen in a repository or in what an agent tells the person.\par
\noindent \textit{[\dots{} a description of the input: the numbered conversation and the commands that a replay cannot run \dots{}]}\par
\smallskip
\noindent \textbf{How to read it}\par
\noindent Read every message of the person's first, in order, before reading what the agent did. The person's messages are what the target is written from.\par
\smallskip
\noindent \textbf{What becomes a clause}\par
\noindent A clause exists for one of two reasons, and you say which:\par
\hangindent=2em\hangafter=1\noindent\hspace*{1em}\makebox[1em][l]{\textbullet}\textbf{Asked.} A message of the person's asks for it. Cite that message.\par
\hangindent=2em\hangafter=1\noindent\hspace*{1em}\makebox[1em][l]{\textbullet}\textbf{Required.} Something the person asked for cannot hold without it. Cite the message that asked for that thing, and say in one sentence why it cannot hold without this. A reported failure asks for it to be fixed. \textit{[\dots{}]}\par
\noindent Everything else the recorded agent did is left out, even when the person accepted it: running the checks, adding tests, formatting, committing, refactoring, updating documentation. Seven more rules decide what a clause says:\par
\smallskip
\hangindent=2em\hangafter=1\noindent\hspace*{1em}\makebox[1em][l]{\textbullet}\textbf{Name the end result, never a step on the way.}\par
\hangindent=2em\hangafter=1\noindent\hspace*{1em}\makebox[1em][l]{\textbullet}\textbf{A step the person refused never enters the target.}\par
\hangindent=2em\hangafter=1\noindent\hspace*{1em}\makebox[1em][l]{\textbullet}\textbf{The last word wins.}\par
\hangindent=2em\hangafter=1\noindent\hspace*{1em}\makebox[1em][l]{\textbullet}\textbf{Say what must hold, not how it was built.}\par
\noindent \textit{[\dots{} 3 further rules on values that depend on time, how the recorded session checked a result, and commands that a replay cannot run \dots{}]}\par
\smallskip
\noindent \textbf{Which clause is the main one}\par
\noindent Exactly one clause is \texttt{main}: the result the session was for.\par
\smallskip
\noindent \textbf{What you write}\par
\hangindent=2em\hangafter=1\noindent\hspace*{1em}\makebox[1em][l]{\textbullet}\texttt{clauses}, in the order the person asked for them. For each:\par
\hangindent=3em\hangafter=1\noindent\hspace*{2em}\makebox[1em][l]{\textbullet}\texttt{text}: one result a grader can mark met or not met on its own, in one sentence.\par
\hangindent=3em\hangafter=1\noindent\hspace*{2em}\makebox[1em][l]{\textbullet}\texttt{main}: true for exactly one clause.\par
\hangindent=3em\hangafter=1\noindent\hspace*{2em}\makebox[1em][l]{\textbullet}\texttt{basis}: \texttt{asked} or \texttt{required}.\par
\hangindent=3em\hangafter=1\noindent\hspace*{2em}\makebox[1em][l]{\textbullet}\texttt{asked\_at}: the event numbers of the person's messages that ask for it.\par
\hangindent=3em\hangafter=1\noindent\hspace*{2em}\makebox[1em][l]{\textbullet}\dots{}\par
\hangindent=2em\hangafter=1\noindent\hspace*{1em}\makebox[1em][l]{\textbullet}\texttt{left\_out}: what the recorded agent did or proposed that is not a clause.\par
\hangindent=2em\hangafter=1\noindent\hspace*{1em}\makebox[1em][l]{\textbullet}\dots{}\par
\smallskip
\noindent \raggedright\textbf{[User message]}\par
\noindent the person's messages are at events \texttt{\{user\_message\_event\_numbers\}}\allowbreak{}\par
\smallskip
\noindent \textbf{The conversation}\par
\noindent \texttt{\{rendered\_session\}}\allowbreak{}\par
\smallskip
\noindent \textbf{Programs and commands another run cannot run}\par
\noindent \texttt{\{unavailable\_commands\}}\allowbreak{}\par
\end{tcolorbox}
\caption{Prompt to write the goals of a task from the recorded session alone (GPT-5.6 Terra).
The prompt calls the goals a target, each goal a clause, and the LLM judge of task success a grader.}
\label{fig:prompt-goal-writer}
\end{figure*}

\begin{figure*}[tp]
\begin{tcolorbox}[colback=gray!5!white, colframe=gray!75!black, title=Identifying actions the user pushed back on, fontupper=\small]
\noindent \textbf{[System prompt]}\par
\noindent \textbf{Read one recorded session for what stood between the person and their final outcome}\par
\smallskip
\noindent You read one recorded session between a person and a coding agent, together with the result the person asked for over the session, written as numbered clauses with one marked main (the target). You write three things, in this order: a ledger with one row per block, the agent's behaviours, and the frictions.\par
\smallskip
\noindent Hindsight is allowed and expected: you know the target and how the session ended, and you read every block against them.\par
\smallskip
\noindent \textit{[\dots{} a description of the input \dots{}]}\par
\smallskip
\noindent \textbf{How to read it}\par
\smallskip
\noindent Read the whole session first, end first: how it ended against the target, what was reverted, what was redone, what was kept, and what the person asked late that they could have asked early. Only then go forward block by block.\par
\smallskip
\noindent A \textbf{block} is one message from the person and everything the agent said and did before the person's next message.\par
\smallskip
\noindent \textbf{The ledger}\par
\smallskip
\noindent One row per block, in order, every block included.\par
\smallskip
\hangindent=2em\hangafter=1\noindent\hspace*{1em}\makebox[1em][l]{\textbullet}\texttt{block\_event\_seq}: the event number of the person's message that opened it.\par
\noindent \textit{[\dots{} further fields \dots{}]}\par
\hangindent=2em\hangafter=1\noindent\hspace*{1em}\makebox[1em][l]{\textbullet}\texttt{person\_response}: what the person's next message did about this block's work: \texttt{continued}, \texttt{corrected}, \texttt{objected}, \texttt{questioned}, \texttt{asked\_to\_understand}, \texttt{reverted}, \texttt{repeated} or \texttt{none}. \textit{[\dots{} the meaning of each \dots{}]}\par
\smallskip
\noindent \textbf{The behaviours}\par
\smallskip
\noindent \textbf{One row is one act the agent took or withheld, that one instruction to an agent would change.} \textit{[\dots{}]}\par
\smallskip
\noindent Rows come from two places, and every candidate goes in:\par
\smallskip
\hangindent=2.4em\hangafter=1\noindent\hspace*{1em}\makebox[1.4em][l]{1.}\textbf{Every block whose \texttt{person\_response} is \texttt{corrected}, \texttt{objected}, \texttt{questioned}, \texttt{reverted} or \texttt{repeated}.}\par
\hangindent=2.4em\hangafter=1\noindent\hspace*{1em}\makebox[1.4em][l]{2.}\textbf{Every stretch the ledger marks \texttt{undid} or \texttt{unchanged} that the agent caused}, and any act whose cost shows only later (work redone, a need that arrived late).\par
\smallskip
\noindent \textit{[\dots{} the fields of each row \dots{}]}\par
\smallskip
\noindent \textbf{The frictions}\par
\smallskip
\noindent A friction is what stood between the person and the target over the session: a stretch of blocks where progress stalled or went backwards, work that was later undone or shown unnecessary, or a fact that was needed from the start and arrived late.\par
\smallskip
\noindent Which behaviours are frictions:\par
\smallskip
\hangindent=2.4em\hangafter=1\noindent\hspace*{1em}\makebox[1.4em][l]{1.}\textbf{A behaviour the person pushed back on is a friction}, unless its \texttt{why\_not} says what makes the push-back not a cost against the target.\par
\hangindent=2.4em\hangafter=1\noindent\hspace*{1em}\makebox[1.4em][l]{2.}\textbf{A behaviour nobody reacted to is a friction only when what happened later shows its cost}, such as the work being redone or a need arriving late.\par
\hangindent=2.4em\hangafter=1\noindent\hspace*{1em}\makebox[1.4em][l]{3.}\textbf{Work wider than the person asked for, or work the agent did not check, is a friction only when the person reverted it, objected to it or questioned it.}\par
\hangindent=2.4em\hangafter=1\noindent\hspace*{1em}\makebox[1.4em][l]{4.}\textbf{A slip the agent fixed inside the same block, with no response from the person, is not a friction.}\par
\smallskip
\noindent \textit{[\dots{} the fields of each action \dots{}]}\par
\smallskip
\noindent \textbf{What must never appear in what you write}\par
\smallskip
\noindent No web address, no filesystem path, no file name, no person's name, no product's name, no project's name, and no quoted line of the transcript. \textit{[\dots{}]}\par
\smallskip
\noindent \raggedright\textbf{[User message]}\par
\noindent the person's messages that open the blocks, one ledger row each, in this order: \texttt{\{block\_event\_numbers\}}\allowbreak{}\par
\smallskip
\noindent \textbf{The target}\par
\smallskip
\noindent \texttt{\{numbered\_goals\}}\allowbreak{}\par
\smallskip
\noindent \textbf{What a run environment can do}\par
\smallskip
\noindent \texttt{\{environment\_facts\}}\allowbreak{}\par
\smallskip
\noindent \textbf{The conversation}\par
\smallskip
\noindent \texttt{\{rendered\_session\}}\allowbreak{}\par
\end{tcolorbox}
\caption{Prompt to list the agent actions that the user pushed back on in a recorded session (GPT-5.6 Terra).
The prompt calls each action the user pushed back on a friction, and an agent turn a block.}
\label{fig:prompt-pushback-reader}
\end{figure*}

\clearpage

\begin{figure*}[tp]
\begin{tcolorbox}[colback=gray!5!white, colframe=gray!75!black, title=Writing the policy instructions, fontupper=\small]
\noindent \textbf{[System prompt]}\par
\noindent \textbf{Write the disposition behind each friction}\par
\smallskip
\noindent You are given the result a person asked for in a recorded session with a coding agent, as numbered clauses (the target), how the session ended against that target, what a run environment can do, and the frictions. You do not see the rest of the recording. That is deliberate: you are to write behaviours, not this task's answer.\par
\smallskip
\noindent \textbf{What you write}\par
\smallskip
\noindent For every friction whose \texttt{kinds} is not \texttt{["none"]} and whose \texttt{agent\_contributed} is \texttt{yes}, two sentences describing two agents in the same situation: \texttt{observed} and \texttt{suited}. Every sentence is an instruction \textbf{to} a coding agent that has tools for reading, searching, editing and running commands, and it has one fixed shape.\par
\smallskip
\noindent The shape, in order:\par
\smallskip
\hangindent=2.4em\hangafter=1\noindent\hspace*{1em}\makebox[1.4em][l]{1.}\textbf{The situation, as the agent meets it.} Begin with "Whenever" and describe the moment from the agent's own seat. \textit{[\dots{} how to describe the kind of request without its words \dots{}]}\par
\hangindent=2.4em\hangafter=1\noindent\hspace*{1em}\makebox[1.4em][l]{2.}\textbf{The order.} For a behaviour that adds a step: "do this before anything else, before answering and before any edit". For a behaviour that withholds a step: say plainly what is not done and until when.\par
\hangindent=2.4em\hangafter=1\noindent\hspace*{1em}\makebox[1.4em][l]{3.}\textbf{The act, as a tool-visible operation.} Name the operation the agent performs: search the whole repository for X, read the file that Y, run Z, ask the person W. \textbf{The act must be one the run environment can perform.} \textit{[\dots{} what to do when avoiding the action needs something outside the starting environment \dots{}]}\par
\hangindent=2.4em\hangafter=1\noindent\hspace*{1em}\makebox[1.4em][l]{4.}\textbf{Where the result goes.} Say where in the reply the act's result appears.\par
\hangindent=2.4em\hangafter=1\noindent\hspace*{1em}\makebox[1.4em][l]{5.}\textbf{The control clause}, when the friction involves one: whether to act or to wait, and until what.\par
\smallskip
\noindent Steps 2 to 5 shape the suited sentence. The observed sentence begins with the same situation (step 1) and then states only the acts the recorded agent actually took, in the order it took them. The observed and suited sentences differ only in the acts and their order, never in the situation.\par
\smallskip
\noindent \textit{[\dots{} rules that keep each sentence to its situation \dots{}]}\par
\smallskip
\noindent A sentence is a standing behaviour, not a plan for this task. The test: the same sentence could be handed to an agent on a different task, in a different repository, unchanged, and still tell it what to do.\par
\smallskip
\noindent The suited sentence may be no stronger than the evidence.\par
\smallskip
\noindent For each friction:\par
\hangindent=2em\hangafter=1\noindent\hspace*{1em}\makebox[1em][l]{\textbullet}\texttt{friction\_index}: which friction it is written from, counting from 1, as given.\par
\hangindent=2em\hangafter=1\noindent\hspace*{1em}\makebox[1em][l]{\textbullet}\texttt{antecedent}: the situation, in a few words.\par
\hangindent=2em\hangafter=1\noindent\hspace*{1em}\makebox[1em][l]{\textbullet}\texttt{observed}, \texttt{suited}: the two sentences.\par
\hangindent=2em\hangafter=1\noindent\hspace*{1em}\makebox[1em][l]{\textbullet}\dots{}\par
\smallskip
\noindent \textit{[\dots{} the actions to skip, and words that must never appear \dots{}]}\par
\smallskip
\noindent \raggedright\textbf{[User message]}\par
\noindent \textbf{The target}\par
\smallskip
\noindent \texttt{\{numbered\_goals\}}\allowbreak{}\par
\smallskip
\noindent \textbf{How the session ended against it}\par
\smallskip
\noindent \texttt{\{outcome\_against\_target\}}\allowbreak{}\par
\smallskip
\noindent \textbf{What a run environment can do}\par
\smallskip
\noindent \texttt{\{environment\_facts\}}\allowbreak{}\par
\smallskip
\noindent \textbf{The frictions}\par
\smallskip
\noindent \texttt{\{frictions\_json\}}\allowbreak{}\par
\end{tcolorbox}
\caption{Prompt to turn each action that the user pushed back on into an instruction for the observed policy and one for the proactive policy (GPT-5.6 Terra).
The prompt calls the proactive instruction the suited one.}
\label{fig:prompt-policy-writer}
\end{figure*}

\begin{figure*}[tp]
\begin{tcolorbox}[colback=gray!5!white, colframe=gray!75!black, title=Judging task success, fontupper=\small]
\noindent \textbf{[System prompt]}\par
\noindent You grade whether the work a coding agent left behind meets each part of the result a user needed.\par
\smallskip
\noindent You receive the result as numbered clauses, one marked main, the difference between the repository as the session began and as it ended (new files in full), and the agent's turns in order: what it said, and each file it edited and command it ran, with the last lines each command printed.\par
\smallskip
\noindent Judge the state at the end of the session. Work that was done and later undone does not count.\par
\smallskip
\noindent Mark every clause on its own, in order, with these rules:\par
\smallskip
\hangindent=2.4em\hangafter=1\noindent\hspace*{1em}\makebox[1.4em][l]{1.}A clause about the code is judged from the repository difference. It is met when the final code does what the clause asks, however it was implemented. The agent's description of its code never stands in for the code.\par
\hangindent=2.4em\hangafter=1\noindent\hspace*{1em}\makebox[1.4em][l]{2.}A clause about a check (tests, lint, formatting, a build, a command working) is met when the last run of that check printed a passing result. \textit{[\dots{}]}\par
\hangindent=2.4em\hangafter=1\noindent\hspace*{1em}\makebox[1.4em][l]{3.}A clause about something the repository cannot show (a push, a pull request, a copy to the clipboard, a program run elsewhere) is met when a command's printed output shows it, or when the agent reports having done it and nothing it printed or said later contradicts the report.\par
\hangindent=2.4em\hangafter=1\noindent\hspace*{1em}\makebox[1.4em][l]{4.}A clause about telling the user something is met when the agent's messages, taken together by the end, give the user every point the clause names.\par
\hangindent=2.4em\hangafter=1\noindent\hspace*{1em}\makebox[1.4em][l]{5.}A clause with several parts, or about all or every case, is met only when every part and every case holds.\par
\hangindent=2.4em\hangafter=1\noindent\hspace*{1em}\makebox[1.4em][l]{6.}A clause applies unless it only makes sense on a way of doing the work that this run did not take. \textit{[\dots{}]}\par
\hangindent=2.4em\hangafter=1\noindent\hspace*{1em}\makebox[1.4em][l]{7.}When the evidence neither shows a clause holds nor shows it fails, it is not met. Say what is missing.\par
\smallskip
\noindent For each clause give its index, then one or two sentences of evidence naming what shows it or what is missing, then the block IDs it rests on, then outside\_the\_run, applies and met. Do not judge the user's messages, the agent's tone, or how many turns it took.\par
\smallskip
\noindent \raggedright\textbf{[User message]}\par
\noindent \textbf{The result the user asked for, one clause per line}\par
\smallskip
\noindent \texttt{\{numbered\_clauses\}}\allowbreak{}\par
\smallskip
\noindent \textbf{What changed in the repository}\par
\smallskip
\noindent \texttt{\{workspace\_difference\}}\allowbreak{}\par
\smallskip
\noindent \textbf{What the agent said, one turn per row}\par
\smallskip
\noindent \texttt{\{agent\_turns\_json\}}\allowbreak{}\par
\end{tcolorbox}
\caption{Prompt to judge task success (GPT-6.1 Sol).}
\label{fig:prompt-grader}
\end{figure*}

\clearpage

\begin{figure*}[tp]
\begin{tcolorbox}[colback=gray!5!white, colframe=gray!75!black, title=Pushback check in the user study, fontupper=\small]
\noindent \textbf{[System prompt]}\par
\noindent You read part of a recorded session between a person and a coding agent. It ends with one exchange in full: the person's message, everything the agent said and did in answer, and the person's next message. You say whether the person's last message pushes back on what the agent did or said, in its last turn or earlier.\par
\smallskip
\noindent \textit{[\dots{} how turns and interruptions are shown \dots{}]}\par
\smallskip
\noindent Pushing back is a reaction against the agent's work or claims. Choose one kind:\par
\smallskip
\hangindent=2em\hangafter=1\noindent\hspace*{1em}\makebox[1em][l]{\textbullet}corrected: the person says the agent got something wrong, misunderstood, or did it differently from what was asked, and wants it fixed.\par
\hangindent=2em\hangafter=1\noindent\hspace*{1em}\makebox[1em][l]{\textbullet}rejected: the person refuses, reverts or declines what the agent did or proposed, including a tool call the person refused to allow.\par
\hangindent=2em\hangafter=1\noindent\hspace*{1em}\makebox[1em][l]{\textbullet}interrupted: the person stopped the agent before it finished its turn.\par
\hangindent=2em\hangafter=1\noindent\hspace*{1em}\makebox[1em][l]{\textbullet}redirected: because of what the agent did, the person changes the approach or what the agent should work on.\par
\hangindent=2em\hangafter=1\noindent\hspace*{1em}\makebox[1em][l]{\textbullet}none: the person does not push back. They accept the work and move on, give the next step after the agent finished what was asked, answer the agent's question, add information the agent could not have had, or ask for information or an explanation.\par
\noindent \textit{[\dots{} an example message for each kind \dots{}]}\par
\smallskip
\noindent A question can push back. A question that doubts a choice the agent made, or points at a case the agent's answer does not cover, is corrected. \textit{[\dots{} examples \dots{}]}\par
\smallskip
\noindent Judge from what the person wrote and what the agent had just done, not from the tone alone.\par
\smallskip
\noindent Answer with \texttt{pushed\_back} (true for any kind but none), \texttt{kind}, and \texttt{reason}:\par
\noindent one sentence saying what in the person's message, against which part of the agent's turn, decides it. Cite event numbers; do not quote.\par
\smallskip
\noindent \raggedright\textbf{[User message]}\par
\noindent The person's message at event \texttt{\{point\_event\_number\}}\allowbreak{} is the one to judge.\par
\smallskip
\noindent \textbf{Earlier messages}\par
\smallskip
\noindent \texttt{\{earlier\_messages\}}\allowbreak{}\par
\smallskip
\noindent \textbf{The last exchange}\par
\smallskip
\noindent \texttt{\{last\_exchange\}}\allowbreak{}\par
\end{tcolorbox}
\caption{Prompt to decide whether the participant's message at a candidate point pushes back on the agent, used in the user study (GPT-5.6 Luna).}
\label{fig:prompt-study-pushback}
\end{figure*}

\clearpage

\end{document}